\documentclass[times,review,10pt]{elsarticle}

\usepackage{amssymb}
\usepackage{framed,multirow}

\usepackage{latexsym}
\usepackage{textcomp,booktabs}
\usepackage{amssymb}
\usepackage{pifont}
\usepackage{url}
\usepackage{cite}
\usepackage{natbib}
\usepackage{xcolor}
\definecolor{newcolor}{rgb}{.8,.349,.1}
\definecolor{green2}{RGB}{52, 159, 43}
\definecolor{SlateBlue}{RGB}{106,90,205} 
\definecolor{DarkRed}{RGB}{178,34,34} 
\usepackage{marvosym}

\usepackage{hyperref} 
\hypersetup{hidelinks, colorlinks=true, linkcolor=red, citecolor=blue, urlcolor=magenta}

\journal{Pattern Recognition}

\begin{document}

\begin{frontmatter}




\title{SequencePAR: Understanding Pedestrian Attributes via A Sequence Generation Paradigm}






\author{Jiandong Jin\textsuperscript{1},  Xiao Wang\textsuperscript{3}\textsuperscript{\Letter}, Yin Lin\textsuperscript{4}, Chenglong Li\textsuperscript{2}, \\ 
        Lili Huang\textsuperscript{3}, Aihua Zheng\textsuperscript{2}, Jin Tang\textsuperscript{3}} 
\address{
1. State Key Laboratory of Opto-Electronic Information Acquisition and Protection Technology, School of Computer Science and Technology, Anhui University, Hefei 230601, China. \\ 
2. State Key Laboratory of Opto-Electronic Information Acquisition and Protection Technology, the School of Artificial Intelligence, Anhui University, Hefei 230601, China. \\
3. Anhui Provincial Key Laboratory of Multimodal Cognitive Computation, School of Computer Science and Technology, Anhui University, Hefei 230601, China. \\
4. iFLYTEK AI Research Institute, Hefei 230088, China. \\
}


\begin{abstract}
Current pedestrian attribute recognition (PAR) algorithms use multi-label or multi-task learning frameworks with specific classification heads. These models often struggle with imbalanced data and noisy samples. Inspired by the success of generative models, we propose Sequence Pedestrian Attribute Recognition (SequencePAR), a novel sequence generation paradigm for PAR. SequencePAR extracts pedestrian features using a language-image pre-trained model and embeds the attribute set into query tokens guided by text prompts. A Transformer decoder generates human attributes by integrating visual features and attribute query tokens. The masked multi-head attention layer in the decoder prevents the model from predicting the next attribute during training. The extensive experiments on multiple PAR datasets validate the effectiveness of SequencePAR.  {Specifically, we achieve 84.92\%, 90.44\%, 90.73\%, and 90.46\% in accuracy, precision, recall, and F1-score on the PETA dataset.}
\end{abstract}

\begin{keyword}
Pedestrian Attribute Recognition, Pre-trained Big Models, Prompt Learning, Image Captioning, Sequence Generation 
\end{keyword}

\end{frontmatter}


\section{Introduction}  
Pedestrian Attribute Recognition (PAR)~\citep{wang2022PARsurvey} involves utilizing a predefined set of attributes to estimate or infer semantic characteristics for pedestrian images, such as \textit{age}, \textit{gender}, \textit{dress}, \textit{movement}, etc. It plays an important role in many practical scenarios like intelligent video monitoring, autonomous driving, and pedestrian analysis. Pedestrian attribute recognition also plays a significant role in other pedestrian-related tasks, such as assisting in matching IDs for pedestrian re-identification tasks~\citep{zhu2023attribute}, object detection~\citep{zhou2019multi}, natural language description-based pedestrian search or retrieval tasks~\citep{peng2023learning}, and analyzing and mining pedestrian attributes to predict pedestrian behavior intentions and possible action trajectories~\citep{li2021uavhuman}.

\begin{figure}
    \centering
    \includegraphics[width=1\linewidth]{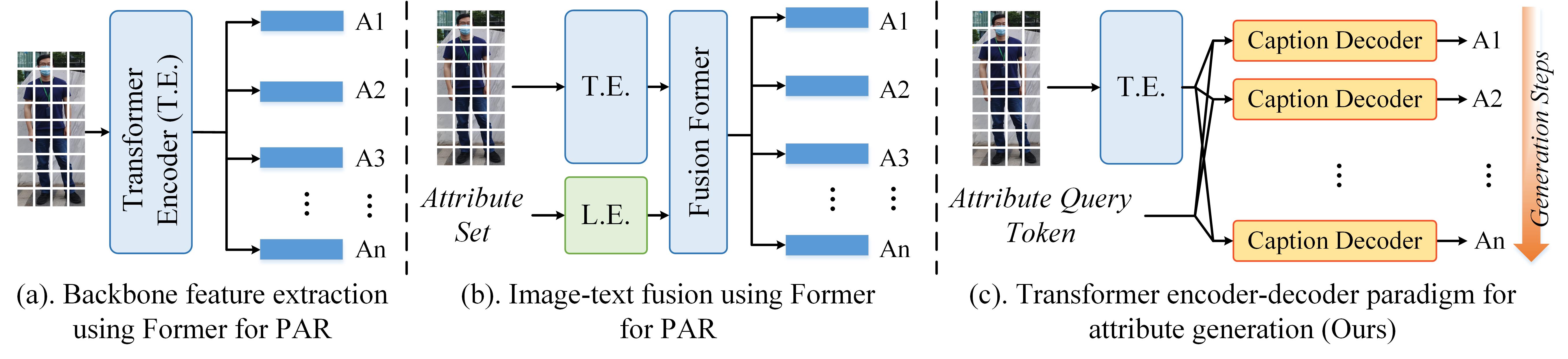}
    \caption{Illustration of different pedestrian attribute recognition frameworks based on Transformer networks. T.E. and L.E. are short for Transformer Encoder and Language Encoder, respectively. Note that existing methods (a, b) follow the discriminative framework, while ours (c) belongs to the generative paradigm.} 
    \label{frontIMG}
\end{figure}

Due to its irreplaceable important role, many deep learning-based pedestrian attribute recognition algorithms have been proposed recently and made great progress. Specifically, numerous researchers have employed Convolutional Neural Networks (CNNs) to extract feature representations from pedestrian images, followed by the use of fully connected layers to regress the attribute responses~\citep{shen2024sspnet}. Additionally, models like Recurrent Neural Networks (RNNs) and Graph Neural Networks (GNNs) have been introduced to capture the complex relationships between human attributes, enhancing the CNN features~\citep{Fan2022CGCN}. More recently, Transformers~\citep{vaswani2017Former}, known for their ability to model long-range dependencies, have been adapted for pedestrian attribute recognition tasks~\citep{cheng2022VTB}. For example, Cheng et al. propose VTB~\citep{cheng2022VTB}, which achieves high performance via multimodal Transformer-based image-text fusion. These advances have undoubtedly propelled the field of pedestrian attribute recognition.

However, despite these advancements, existing pedestrian attribute recognition algorithms typically operate within the multi-label classification framework or multi-task learning, which are categorized as discriminative models. It is well known that discriminative models are susceptible to various challenges, including data imbalance and sparsity, noisy annotations, and insufficient semantic relationship mining, which may constrain their performance.


    
    

To address the limitations of discriminative models, we propose a shift in perspective: ``\emph{{Can we break away from the category of discriminative models and re-examine pedestrian attribute recognition from the perspective of generative models?}}" Generative models, by their nature, can better capture complex relationships and dependencies between attributes. Unlike discriminative models, which focus on classifying or regressing individual attributes independently, generative models transform $N$ independent binary classification problems into a sequence joint probability model $P\left(A_1, A_2,\ldots, A_N\right)=\prod{P\left(A_i\middle|A<i\right)}$, dynamically constructing conditional probability relationships between attributes during the sequence generation process. This explicit modeling approach enables tail attributes to infer from the contextual information of head attributes (e.g., the "skirt" attribute provides strong semantic constraints for predicting "female"), effectively alleviating the issue of poor learning of tail features due to isolated classification in traditional methods. Regarding noise robustness, during the autoregressive generation process, the prediction of each attribute is constrained by the semantic constraints of previously generated attributes. For example, when a noisy sample incorrectly labels "short-sleeve + skirt" as male, the model can automatically correct the subsequent gender prediction through the already generated "skirt" attribute (strongly associated with female). This reasoning mechanism constructs a dynamic error tolerance space. These characteristics make generative models particularly suited for pedestrian attribute recognition, where relationships between attributes play a crucial role in achieving high performance.

Inspired by the aforementioned observations and thinking, in this work, we propose a novel sequence generation paradigm for pedestrian attribute recognition, termed SequencePAR. A comparison between existing Transformer-based pedestrian attribute recognition algorithms and our SequencePAR is illustrated in Figure~\ref{frontIMG}. The key insight of this work is that we formulate attribute recognition as an image captioning task, which can model the relations between human attributes by generating descriptive captions for human images. This generative model allows us to encapsulate the underlying connections between attributes. Furthermore, the burgeoning evidence from various large-scale models strongly indicates the exceptional performance of big language models in precisely this domain. 
In our practical implementation, given the pedestrian image, we first partition it into non-overlapping tokens and adopt the CLIP visual encoder~\citep{radford2021CLIP} to get the visual tokens due to its strong feature expression ability and generalization. We also take the attribute descriptions as the input to capture the high-level semantic information. We adopt the word embedding to transform the attributes into text embeddings and concatenate them with text prompts as the attribute query token. Then, we feed the attribute query tokens into the \emph{masked} multi-head attention layer, and the outputs are fed into the multi-head attention layer together with visual tokens for attribute generation. Note that, the normalization layers and feed-forward layers are also incorporated in the sequence generation decoder network. In the testing phase, the greedy search mechanism is adopted for attribute generation. An overview of our proposed SequencePAR framework can be found in Figure~\ref{framework}.


To sum up, the key contributions of this paper can be summarized as the following three aspects: 

\begin{itemize}
\item [$\bullet$] 
We propose a new generative pedestrian attribute recognition framework, termed SequencePAR, which broke away from the fixed mindset of multi-label classification. It is the first work to handle PAR in a sequence generation manner which handles the issue of imbalanced and noisy attribute learning to some extent.

\item [$\bullet$] 
We propose a novel masked Transformer decoder that predicts each attribute sequentially based on pedestrian tokens and textual representations. It addresses the issue of weak connection of attribute context in the standard multi-label classification.

\item [$\bullet$] 
Extensive experiments on multiple popular pedestrian attribute recognition datasets fully validated the effectiveness of our proposed SequencePAR. The source code and pre-trained models will be released at \url{https://github.com/Event-AHU/OpenPAR}.
\end{itemize}

The rest of this paper is organized as follows: We review related works most related to our paper in Section~\ref{relatedWorks}. The approach we proposed is mainly introduced in Section~\ref{Methodology}, including the problem formulation, overview, input representation, sequence generation module, and loss function. In Section~\ref{experiments}, we conduct extensive experiments and give an in-depth analysis of multiple benchmark datasets. We conclude this paper and propose possible research directions in Section~\ref{conclusion}.

\section{Related Works} \label{relatedWorks}

In this section, we will give a brief introduction to the related works on Pedestrian Attribute Recognition, Transformer Networks, and Sequence Generation. More related works can be found in the following survey~\citep{wang2022PARsurvey} \footnote{\url{github.com/wangxiao5791509/Pedestrian-Attribute-Recognition-Paper-List}}.   
 
\subsection{Pedestrian Attribute Recognition~}   
Pedestrian attribute recognition methods can be categorized into various approaches, such as CNN, RNN-based algorithms, attention mechanisms, and transformer-based methods. Early CNN-based approaches have demonstrated promising results in pedestrian attribute recognition. For instance, 

DAFL~\citep{jia2022learning} enables the model to learn attribute-independent visual features through a group-shared attention library, thereby enhancing the discriminative capability of the predictions.  Shen et al.~\citep{shen2024sspnet} proposed a method that leverages multi-scale priors and attribute-space priors to enable the model to accurately localize attribute regions. Zhou et al.~\citep{zhou2025solution} proposed extracting attribute-specific cues to mitigate the model’s over-reliance on attribute co-occurrence patterns in the dataset, thereby enabling the model to better adapt to imbalanced distributions.
The RNN model is introduced into the PAR community to model the semantic association between human attributes. By incorporating the previously predicted labels, visual features can be dynamically adjusted for the subsequent moments. GRL~\citep{zhao2018GRL} employs an RNN to capture the relationships and mutual exclusivities among the attributes.
The attention mechanism is widely used in deep neural networks and attribute recognition tasks. For example, 
Liu et al. introduce HydraPlus-Net~\citep{2017pa100k}, an architecture that leverages the multi-level feature maps to capture diverse pedestrian details and extract more comprehensive features. 

Inspired by the success of Transformer networks in natural language processing and computer vision tasks, some researchers also adapted Transformers~\citep{vaswani2017Former, dosovitskiy2020VIT, kenton2019bert} for pedestrian attribute recognition. For instance, Wu et al.~\citep{wu2025rethinking} introduce perturbations into the attention maps during training to reduce the model’s focus on non-attribute regions, thereby enhancing its ability to capture attribute-specific areas. Cheng et al.~\citep{cheng2022VTB} approached pedestrian attribute recognition as a visual language task by converting the attribute list into text features. They fed both the visual features and the text features into a Transformer model, facilitating effective interaction between the modalities. 

However, discriminative models face various challenges that may constrain their performance:
\begin{itemize}

    \item \textbf{Data Imbalance and Sparsity}: Models that directly use CNNs or Transformers to predict attribute responses often struggle with data imbalance, where a large number of negative samples can lead to sparse attribute predictions. This is particularly problematic in pedestrian attribute recognition, where certain attributes are underrepresented in the dataset.
    
    \item \textbf{Noisy Annotations}: Pedestrian attribute annotation is a complex and error-prone task. Some datasets~\citep{2016rapv1} contain uncertain or noisy attribute labels, which can significantly impact model reliability and performance. 
    
    \item \textbf{Weak Semantic Relationships}: In discriminative models, where attributes are regressed independently, the semantic relationships between different attributes are often weak or underexplored, leading to suboptimal modeling of attribute dependencies.
\end{itemize}
To address these issues, we employ a generative framework that formulates pedestrian attribute recognition as a sequence generation problem. Through iterative attribute generation, our approach explicitly models the semantic relationships between attributes. This addresses the limitations of traditional pedestrian attribute recognition methods, which often overlook the modeling of relationships within attributes or rely on manual assumptions.

\subsection{Transformer Network} 
Based on self-attention mechanisms, the Transformer~\citep{vaswani2017Former} network first caused a significant stir in the field of natural language processing, demonstrating performance far superior to RNN$/$LSTM series algorithms across various text-related tasks. Then, the pre-training techniques developed based on the Transformer further demonstrate its effectiveness, for example, the BERT~\citep{kenton2019bert}. The remarkable success of the Transformer in natural language processing (NLP) has inspired researchers to explore its application in computer vision. Significant efforts have been made to introduce the Transformer into multiple fields. For example, the ViT~\citep{dosovitskiy2020VIT} proposed by Dosovitskiy et al. is the first milestone of the Transformer-based foundation model. It splits the input image into token representations and introduces a class token before feeding it into the Transformer layers for classification. M. Cornia et al.~\citep{2020m2} are pioneers in applying the Transformer model to the image captioning task.

After that, the Transformer network is also introduced into the multi-modal community, especially for the vision-language-based pre-training and downstream tasks. Many representative vision-language models are proposed one after another, such as the CLIP~\citep{radford2021CLIP}. Inspired by these works, we adopt the Transformer-based pre-trained models to extract the visual and textual representations. More importantly, we formulate the pedestrian attribute recognition task as a sequential text generation problem to better capture the semantic relations between various attributes.

\subsection{Sequence Generation Models~} 
Many high-level tasks involve generating sequential outputs, such as machine translation, image captioning, and speech recognition. These models usually follow the encoder-decoder framework and achieve this target via recurrent neural networks (RNN) or Transformer decoders. More in detail, the LSTM is widely used for temporal information processing. 
Chen et al.~\citep{chen2023seqtrack} formulate the visual tracking as a sequence generation problem, which predicts object bounding boxes in an auto-regressive fashion. 
Li et al.~\citep{li2023blip2} propose the BLIP-2 which bootstraps vision-language pre-training using a lightweight Querying Transformer. This work demonstrates that image-to-text generation can follow the introductions of natural language. 
Peng et al.~\citep{peng2025unbiased} propose an unbiased VQA framework that highlights the potential of generative formulations in visual question answering, further demonstrating the effectiveness of sequence generation paradigms in multimodal reasoning tasks. 
In this work, we formulate pedestrian attribute recognition as a sequence generation problem instead of the standard multi-label classification problem.

\section{Methodology} \label{Methodology} 
In this section, we will first give an introduction to the existing PAR frameworks and our newly proposed ones. Then, we will introduce the overview of our proposed SequencePAR framework and the input representations of our model, including pedestrian images and attribute phrases. The vision and text encoding process will be provided in detail. After that, we focus on the attribute generation using the Transformer decoder network. Finally, we will introduce the loss function used in our training phase.

\subsection{Problem Formulation}   
Pedestrian attribute recognition targets predicting human attributes $\mathcal{A}_i$, where $i = \{1, 2, ..., N\}$, from a pre-defined attribute set $\mathcal{A}_{set} = \{\mathcal{A}_1, \mathcal{A}_2, ..., \mathcal{A}_M\}$, $N \leq M$, based on a given pedestrian image $\mathcal{I}$. Existing deep learning-based attribute recognition models usually treat this task as a multi-label classification or multi-task learning problem~\citep{cheng2022VTB, fan2023parformer}. They usually adopt a backbone network to extract the deep features of the human image and learn to recognize the attributes using multi-category classifiers (MCC), i.e., 
\begin{equation}
    \label{mutlilabelClassification} 
    \mathcal{A}_i = MCC(VB(\mathcal{I})),  
\end{equation}
where $VB$ denotes the visual backbone. 
Some researchers have proposed recognizing human attributes by combining pedestrian image features and attribute text representations. 
\begin{equation}
    \label{imgTextFusionPAR} 
    \mathcal{A}_i = MCC(MMFormer(VB(\mathcal{I}), TB(\mathcal{A}_{set}))),  
\end{equation}
where $MMFormer$ denotes the multi-modal Transformer, $TB$ and $\mathcal{A}_{set}$ denotes the text backbone and raw attribute set, respectively. 
Different from previous works, in this paper, we propose a new sequence generation framework for pedestrian attribute recognition, 
\begin{equation}
    \label{SGFramework} 
    \mathcal{A}_i = SGNet(VB(\mathcal{I}), TB(\mathcal{A}_{set})), 
\end{equation}
where $SGNet$ is short for sequence generation network.

\begin{figure*}
\centering
\small
\includegraphics[width=\linewidth]{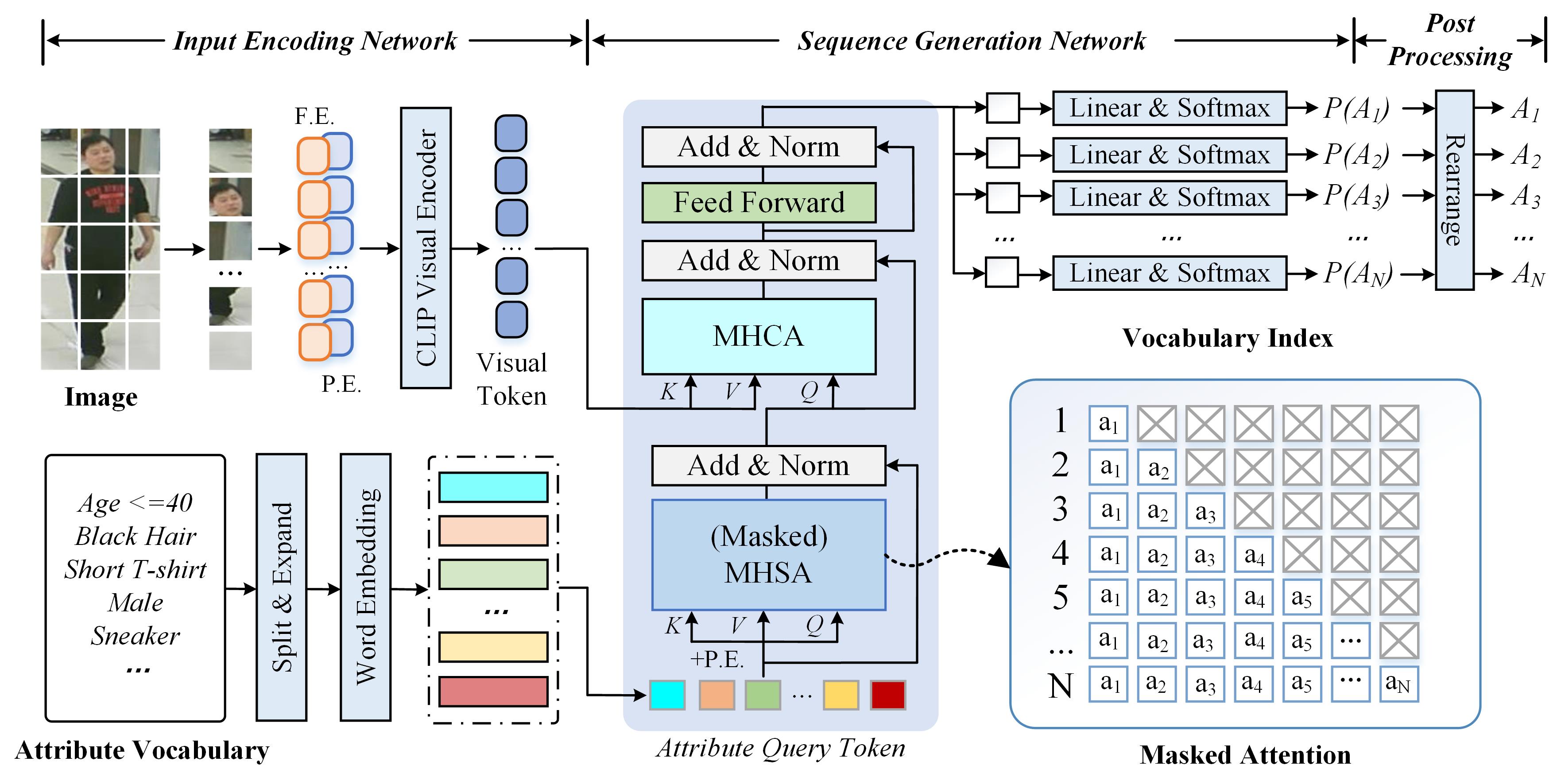}
\caption{\textbf{An overview of our proposed SequencePAR framework.}
Given a pedestrian image, we first partition it into non-overlapping patches and extract visual tokens using pre-trained large models. Unlike previous approaches that directly regress attribute scores through dense layers, our method introduces a novel generative paradigm that sequentially generates attribute descriptions.
}
\label{framework}
\end{figure*}

\subsection{Overview} 
In this work, we formulate the pedestrian attribute recognition task as a sequence generation problem similar to image captioning and machine translation. Given the pedestrian image, we first partition and project it into token representations as the input of the Transformer network. To capture the high-level semantic information of attributes, we also take the raw attributes as the input and utilize the embedded attribute features as the attribute query tokens. The attribute query tokens are processed by masked self-attention layers and further fused with pedestrian image features via cross-attention layers. Such a joint processing of image and text features facilitates the effective integration of multi-modal data for attribute prediction. Finally, the resulting features pass through a linear layer followed by a Softmax function to generate predicted probabilities for each attribute. Therefore, we can get the final attributes via rearrangement. An overview of our proposed SequencePAR framework can be found in Figure~\ref{framework}. In the following subsections, we will focus on each detailed procedure to help the readers better understand our proposed SequencePAR model.

\subsection{Network Architecture}

\noindent 
\textbf{Image Encoding Network.~}  
Given a pedestrian attribute recognition dataset $\mathcal{D} = \{(X_i, A_i) | i = \{0, 1, 2, ..., N\}\}$, where $X_i$ and $A_i = \{a_i^1, a_i^2, ..., a_i^M\}$ represents pedestrian images and corresponding attribute labels, respectively. $N$ is the number of pedestrian images in the dataset and $M$ represents the number of labeled attributes. We first split the image into $256$ non-overlapping patches $P_i \in \mathbb{R}^{14\times14}$, then, project them into token representation $T_i \in \mathbb{R}^{256\times1024}$ using one convolutional layer (kernel size 14 $\times$ 14). Following the standard vision Transformer networks, we also introduce the position encoding $PE \in \mathbb{R}^{257\times1024}$ to capture the spatial information of split patches. The token features and position features are added together as the input of the vision Transformer network. In this paper, we adopt the pre-trained vision encoder of multi-modal big model CLIP~\citep{radford2021CLIP} (ViT-L$/$14 is adopted) to extract the feature representations $\hat{T}_I \in \mathbb{R}^{257 \times 768}$. 
The core components of ViT-L$/$14 are the Transformer layers and each layer consists of multi-head self-attention layers, normalization layers, feed-forward layers, and skip connections. Among them, the detailed procedure of multi-head self-attention is the self-attention operation, which can be written as: 
\begin{equation}
\label{selfattention} 
SA(Q, K, V) = Softmax(\frac{QK^T}{\sqrt{d}}V),  
\end{equation}
where $T$ denotes the transpose operation, $d$ is the dimension of input tokens.

\noindent 
\textbf{Attribute Encoding Network.~}
To help the models better understand the pedestrian attributes that need to be classified, in this work, we take the raw attribute phases $\mathcal{A}_{set} = \{\mathcal{A}_1, \mathcal{A}_2, ..., \mathcal{A}_M\}$ as the input. To be specific, we expand and split each attribute phase $\mathcal{A}_i$ into text representations with the help of prompt engineering. For example, the attribute \textit{"age $\leq$ 40"} is firstly processed to \textit{"age less than 40"}, then, transformed into a sentence \textit{"the \underline{age} of this pedestrian is \underline{less than 40} years old"}. After that, we obtain the text tokens $\mathcal{E}_{set} = \{\mathcal{E}_1, \mathcal{E}_2, ..., \mathcal{E}_M\}$ using word embedding. Then, we adopt the text encoder of CLIP model~\citep{radford2021CLIP} to learn the high-level semantic features of human attributes. Therefore, we can get the attribute query tokens $\{\mathcal{Q}_1, \mathcal{Q}_2, ..., \mathcal{Q}_M\}$, which learns the concept of each attribute well. Due to the strong mutual exclusivity or co-occurrence relationships among many attributes, for instance, in the PETA dataset, attributes such as \textit{casual top}, \textit{formal top}, \textit{casual trousers}, \textit{formal trousers}, and \textit{leather shoes} exhibit significant interdependencies. These relationships are difficult to capture using visual features alone. Modeling such dependencies helps the system avoid inconsistent predictions, such as simultaneously predicting \textit{casual top} and \textit{formal trousers}. To address this, we encode attribute descriptions using the CLIP text encoder as queries, enabling the model to learn semantic dependencies among attributes through causal masking during training.

\noindent 
\textbf{Sequence Generation Network.~} 
In this work, we formulate the pedestrian attribute recognition task as a novel sequence generation problem. Given the attribute query tokens $\mathcal{Q}_i, i \in \{1, 2, ..., M\}$, these represent text embeddings obtained by processing attribute words through prompt templates to generate complete sentences, which are then encoded by our text encoder. We use these embeddings as attribute query tokens within our framework. We first introduce the position encoding tokens and add these tokens as the input of the sequence generation network. Specifically, the query tokens are treated as Key (K), Value (V), and Query (Q) and feed into the \textit{masked multi-head attention} module. {This refers to our implementation of attention masking in the self-attention mechanism of the decoder. This method effectively prevents the model from accessing future information during autoregressive predictions.} It receives the predicted attributes from the preceding block and adopts the causal mask to ensure that the predicted pedestrian attributes only depend on its previous sequence cues. As shown in Figure~\ref{framework}, the output attribute $\mathcal{A}_i$ only attends to the attributes less than $i$, with the guidance of the attention mask. Then, the normalization layers and skip connections are adapted to process the obtained features and the output will be treated as the query input of the subsequent multi-head self-attention layer. The visual tokens are used as the key and value inputs. Then, the normalization and feed-forward layers are utilized to further enhance the decoded features $U$.

\noindent 
\textbf{Post-Processing.~}  
After obtaining the decoded features, we propose to map them into a probability distribution of all attributes defined in each attribute recognition dataset using \textit{linear layer} and \textit{Softmax} operator, i.e., 
\begin{equation}
    \label{} 
    [p_1, p_2, ... , p_M] = Softmax(Linear(U)). 
\end{equation}
Then, we transform the index information into its corresponding pedestrian attributes and get the final predictions using the re-arrangement operation. For example, we remove the repeated attributes and special padding tokens.

\noindent 
\textbf{Loss Function.~} 
The Negative Log Likelihood (NLL) loss function is adopted to optimize our network which can be written as: 
\begin{equation}
\label{NLLLoss} 
Loss = - \frac{1}{N}\sum_{i=1}^N(\sum_{j=1}^M w_j * y_{ij} * p_{ij})
\end{equation} 
where $N$ and $M$ represent the number of input samples and attributes, respectively. $y_{ij}$ is the label of the corresponding attribute of the sample, $p_{ij}$ is the value of the linear layer output after log\_softmax, and $w_j = e^{|y_{ij}-r_j|}$ is the sampling weight corresponding to the class used for balancing, $r_j$ represents the class sampling weight.

\subsection{Training and Inference}   
\begin{figure*}
\centering
\small
\includegraphics[width=\linewidth]{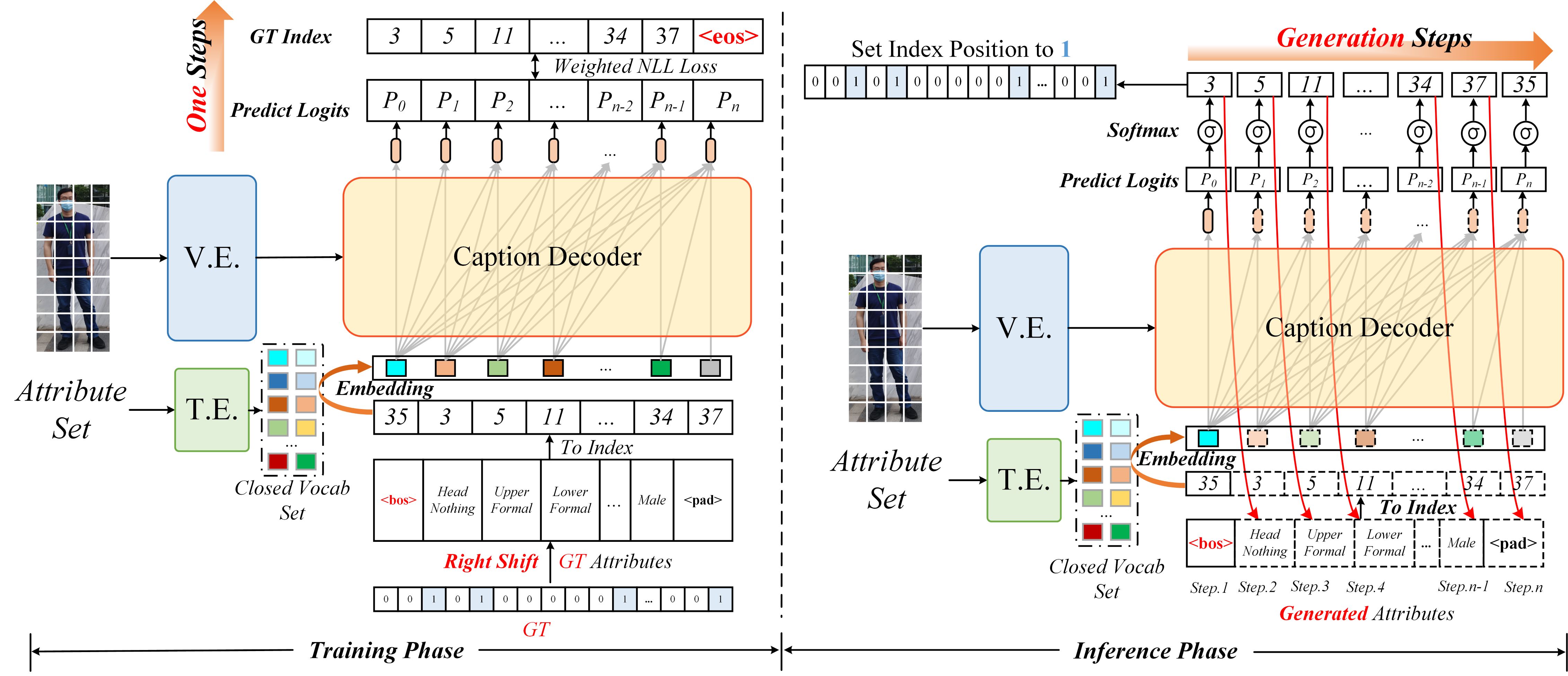}
\caption{The training and inference pipeline of the proposed SequencePAR framework.}
\label{traintest}
\end{figure*}

As shown in Figure~\ref{traintest}, we formulate pedestrian attribute recognition as a closed-set vocabulary sequence generation problem, where the vocabulary consists of all predefined attribute labels along with special tokens, such as the start-of-sequence $<$BOS$>$ and end-of-sequence $<$EOS$>$ markers.
During training, the ground-truth attribute labels corresponding to each image are first converted into textual embeddings using a predefined prompt template. These embeddings are then right-shifted and fed into a Transformer decoder as input queries. Simultaneously, the visual features extracted by a CLIP encoder serve as conditional context for the decoder. The model is trained to predict the probability distribution of the next token in an autoregressive manner at each decoding step.
During inference, the decoder receives the start token and the image’s visual context as input and generates a sequence of attribute indices step-by-step via greedy decoding. At each step, the model uses the previously generated index (converted into an embedding) together with the visual features to predict the next index. The resulting sequence of attribute indices is mapped into a one-hot vector representing the predicted attribute set. Special tokens are removed during post-processing, and duplicated indices are merged to form the final prediction.

\section{Experiments} \label{experiments}
In this section, we will first introduce the datasets and evaluation metric in subsection~\ref{datasetsMetric}. The implementation details are given in subsection~\ref{impleDetails}. After that, we report our recognition results and compare them with other state-of-the-art algorithms in subsection~\ref{CompSOTA}. Then, we conduct extensive studies on our newly proposed SequencePAR framework in subsection~\ref{ablationStudy}. The visualization is also provided in subsection~\ref{visualization} to help the readers better understand our model. Then, we discuss the difference between our model and existing recurrent neural network-based PAR models and also the limitation analysis in subsection~\ref{differenceRNN}, respectively. 

\subsection{Datasets and Evaluation Metric} \label{datasetsMetric}
In this paper, we evaluate our framework on six pedestrian attribute datasets, including \textbf{PETA}, \textbf{RAPv1}, \textbf{RAPv2}, \textbf{PA100K}, \textbf{PETA-ZS}, and \textbf{RAP-ZS}. Note that, the last two datasets are split based on zero-shot setting, i.e., no pedestrian images share the same identity in the training and testing subset. A brief introduction to these datasets is given below.

$-$ \textbf{PETA dataset}~\citep{deng2014peta} contains 19,000 outdoor or indoor pedestrian images and 61 binary attributes. These images are split into training, validation, and testing subsets, which contains 9500, 1900, and 7600 images, respectively. In our experiments, we select 35 pedestrian attributes by following the work~\citep{deng2014peta}.

$-$ \textbf{RAPv1 dataset}~\citep{2016rapv1} contains 41,585 pedestrian images and 69 binary attributes, where 33,268 images are used for training. Usually, in current pedestrian attribute recognition algorithms, 51 attributes are selected for training and evaluation.

$-$ \textbf{RAPv2 dataset}~\citep{2019rapv2} has 84,928 pedestrian images and 69 binary attributes, where 67,943 were used for training. We select 54 attributes for the training and evaluation of our model.

$-$ \textbf{PA100K dataset}~\citep{2017pa100k} is the largest pedestrian attribute recognition dataset which contains 100,000 pedestrian images, and 26 binary attributes. In our experiments, we split them into a training and validation set which contains 90,000 images, and a testing subset with the remaining 10,000 images.

$-$ \textbf{PETA-ZS dataset} is proposed by Jia et al. based on PETA~\citep{deng2014peta} dataset by following the zero-shot protocol. The training, validation, and testing subset contains 11241, 3826, and 3933 samples. 35 common attributes are adopted for our experiments by following Jia et al.~\citep{2021Rethinking}. 

$-$ \textbf{RAP-ZS dataset} is constructed based on RAPv2 and consists of 17,062 images for training, 4,628 for validation, and 4,928 for testing. Importantly, there is no overlap in pedestrian identities between the training and inference sets, ensuring a strict zero-shot evaluation protocol. In our experiments, we follow Jia et al.~\citep{2021Rethinking} and select 53 attributes for evaluation.

For the evaluation metric, the Accuracy, Precision, Recall, and F1-measure are adopted for the experimental comparison. Specifically, the instance-based evaluation metric Accuracy can be expressed as: 
\begin{equation}\label{acc}
Accuracy = \frac{TP+TN}{TP+TN+FP+FN}, 
\end{equation}
where \textit{TP} is predicting the correct positive sample, \textit{TN} is predicting the correct negative sample, \textit{FP} is a negative sample of prediction errors, and \textit{FN} is a positive sample of prediction errors.
The formulation of Precision, Recall, and F1 measures can be expressed as: 
\begin{equation}\label{prec}
Precision=\frac{TP}{TP+FP}, ~~~~ Recall=\frac{TP}{TP+FN}, 
\end{equation}
\begin{equation}\label{f1}
F1 =\frac{2 \times Precision \times Recall}{Precision+Recall}. 
\end{equation}

\begin{table*}[!htb]
\center
\small   
\scriptsize 
\caption{Comparison with state-of-the-art methods on PETA and PA100K datasets. The \textcolor{red}{first} and \textcolor{blue}{second} are shown in \textcolor{red}{red} and \textcolor{blue}{blue}, respectively. "-" means this indicator is not available.} \label{PETAPA100Kresults} 
\resizebox{1\columnwidth}{!}{
\begin{tabular}{l|c|c|cccc|cccc}
\hline \toprule [0.5 pt] 
\multicolumn{1}{c|}{\multirow{2}{*}{Methods}} & \multicolumn{1}{c|}{\multirow{2}{*}{Ref}} & \multicolumn{1}{c|}{\multirow{2}{*}{Backbone}} & \multicolumn{4}{c|}{PETA} & \multicolumn{4}{c}{PA100K} \\ \cline{4-11}
\multicolumn{1}{c|}{} &
  \multicolumn{1}{c|}{} &
  \multicolumn{1}{c|}{} &
  \multicolumn{1}{c}{Accuracy} & 
  \multicolumn{1}{c}{Precision} &
  \multicolumn{1}{c}{Recall} &
  \multicolumn{1}{c|}{F1} &
  \multicolumn{1}{c}{Accuracy} &
  \multicolumn{1}{c}{Precision} &
  \multicolumn{1}{c}{Recall} &
  \multicolumn{1}{c}{F1} \\ \hline
DeepMAR \citep{deepmar} & ACPR 2015 & CaffeNet  & 75.07 & 83.68 & 83.14 & 83.41 & 70.39 & 82.24 & 80.42 & 81.32 \\
HPNet \citep{2017pa100k} & ICCV 2017  & Inception  & 76.13 & 84.92 & 83.24 & 84.07 & 72.19 & 82.97 & 82.09 & 82.53 \\
GRL  \citep{zhao2018GRL} & IJCAI 2018 & Inception-V3  & - & 84.34 & 88.82 & 86.51  & - & - & - & - \\
VAC  \citep{2019VAC} & CVPR 2019 & ResNet50   & - & - & - & -  & 79.44 & 88.97 & 86.26 & 87.59 \\
SSCsoft  \citep{2021ssc} &  ICCV 2021& ResNet50  & 78.95 & 86.02 & 87.12 & 86.99 & 78.89 & 85.98 & 89.10 & 86.87 \\	
DRFormer  \citep{2022drformer} & NC 2022 & ViT-B$/$16  & 81.30 & 85.68 & \textcolor{blue}{\textbf{91.08}} & 88.30 & 80.27 & 87.60 & 88.49 & 88.04 \\
VAC-Combine \citep{guo2022visual} & IJCV 2022  & ResNet50   & - & - & - & - & 80.66 & 88.72 & 88.10 & 88.41 \\
DAFL  \citep{jia2022learning} & AAAI 2022 & ResNet50  & 78.88 & 85.78 & 87.03 & 86.40  & 80.13 & 87.01 & 89.19 & 88.09 \\
CGCN \citep{Fan2022CGCN} & TMM 2022  & ResNet50  & 79.30 & 83.97 & 89.38 & 86.59 & - & - & - & - \\ 
VTB \citep{cheng2022VTB} & TCSVT 2022  & ViT-B$/$16  & 79.60 & 86.76 & 87.17 & 86.71  & 80.89 & 87.88 & 89.30 & 88.21 \\
PARFormer-L \citep{fan2023parformer}  & TCSVT 2023 & Swin-L & 82.86 & 88.06 & \textcolor{red}{\textbf{91.98}} & 89.06 & 81.13 & 88.09 & \textcolor{red}{\textbf{91.67}} & 88.52 \\
DFDT \citep{zheng2023diverse} & EAAI 2023 & Swin-B & 81.17  & 87.44 & 88.96 & 88.19 & 81.24 & 88.02 & 89.48 & 88.74 \\
OAGCN \citep{lu2023orientation} & TMM 2023 & Swin-B & \textcolor{blue}{\textbf{82.95}} & 88.26 & 89.10 & 88.68 & 80.38 & 84.55 & 90.42 & 87.39 \\
VTB* \citep{cheng2022VTB} & TCSVT 2022  & ViT-L$/$14  & 79.59 & 86.66 & 87.82 & 86.97  & 81.76 & 87.87 & \textcolor{blue}{\textbf{90.67}} &  88.86 \\	
{SSPNet}~\citep{shen2024sspnet} & {PR 2024} & Swin-S & 82.80 & 88.48 & 90.55 & \textcolor{blue}{\textbf{89.50}} & 80.63 & 87.79 & 89.32 & 88.55  \\
{ViT-RE++}~\citep{tan2024vision} & {TMM 2024} & ViT-B & 81.64 & 88.59 & 88.82 & 88.70 & 81.47 & \textcolor{blue}{\textbf{89.78}} & 89.77 & 88.88 \\
{SOFA}~\citep{wu2024selective} & {AAAI 2024} & ViT-B & 81.06 & 87.77 & 88.35 & 87.83 & 81.14 & 88.39 & 88.98 & 88.34 \\
{HDFL}~\citep{wu2025rethinking} & {NN 2025} & ViT-B & 79.66 & 87.08 & 87.16 & 86.85 & 80.23 & 87.45 & 88.74 & 87.72 \\
{AAR}~\citep{wu2025high} & {NC 2025} & ViT-B & 82.46 & \textcolor{blue}{\textbf{89.04}} & 89.01 & 88.81 & \textcolor{blue}{\textbf{81.96}} & 88.78 & 89.62 & 88.87 \\
{SOFAFormer++}~\citep{wu2025learning} & {TCSVT 2025} & ViT-B & 80.91 & 87.17 & 88.74 & 87.70 & 80.80 & 87.62 & 89.37 & 88.15 \\
{PIL}~\citep{zhou2025solution} & {IJCV 2025} & ConvNeXt-base & - & - & - & 87.49 & - & - & - & \textcolor{blue}{\textbf{89.53}} \\
\hline
SequencePAR & - & ViT-L$/$14 & \textcolor{red}{\textbf{84.92}} & \textcolor{red}{\textbf{90.44}} & 90.73 & \textcolor{red}{\textbf{90.46}} & \textcolor{red}{\textbf{83.94}} & \textcolor{red}{\textbf{90.38}} & 90.23 & \textcolor{red}{\textbf{90.10}}\\
\hline \toprule [0.5 pt] 
 \end{tabular} } 
\end{table*}

\begin{table*}[!htb]
\center
\small  
\scriptsize 
\caption{Comparison with state-of-the-art methods on RAPv1 and RAPv2 datasets. The \textcolor{red}{first} and \textcolor{blue}{second} are shown in \textcolor{red}{red} and \textcolor{blue}{blue}, respectively. "-" means this indicator is not available.} \label{RAPResult} 
\resizebox{1\columnwidth}{!}{
\begin{tabular}{l|c|c|cccc|cccc}
\hline \toprule [0.5 pt] 
\multicolumn{1}{c|}{\multirow{2}{*}{Methods}} & \multicolumn{1}{c|}{\multirow{2}{*}{Ref}} & \multicolumn{1}{c|}{\multirow{2}{*}{Backbone}} & \multicolumn{4}{c|}{RAPv1} & \multicolumn{4}{c}{RAPv2} \\ \cline{4-11}
\multicolumn{1}{c|}{} &
  \multicolumn{1}{c|}{} &
  \multicolumn{1}{c|}{} &
  \multicolumn{1}{c}{Accuracy} & 
  \multicolumn{1}{c}{Precision} &
  \multicolumn{1}{c}{Recall} &
  \multicolumn{1}{c|}{F1} &
  \multicolumn{1}{c}{Accuracy} &
  \multicolumn{1}{c}{Precision} &
  \multicolumn{1}{c}{Recall} &
  \multicolumn{1}{c}{F1} \\ \hline
DeepMAR \citep{deepmar} & ACPR 2015 & CaffeNet   & 62.02 & 74.92 & 76.21 & 75.56 & - & - & - & - \\
HPNet \citep{2017pa100k} & ICCV 2017  & Inception  & 65.39 & 77.33 & 78.79 & 78.05 & - & - & - & -  \\
GRL  \citep{zhao2018GRL} & IJCAI 2018 & Inception-V3  & - & 77.70 & 80.90 & 79.29 & - & - & - & - \\
VAC  \citep{2019VAC} & CVPR 2019 & ResNet50   & - & - & - & - & 64.51 & 75.77 & 79.43 & 77.10 \\
DRFormer  \citep{2022drformer} & NC 2022 & ViT-B$/$16  & 70.60 & 80.12 & 82.77 & 81.42 & - & - & - & - \\
VAC-Combine \citep{guo2022visual} & IJCV 2022  & ResNet50  & 70.12 & \textcolor{blue}{\textbf{81.56}} & 81.51 & 81.54 & - & - & - & - \\
DAFL  \citep{jia2022learning} & AAAI 2022 & ResNet50   & 68.18 & 77.41 & 83.39 & 80.29 & 66.70 & 76.39 & 82.07 & 79.13 \\
CGCN \citep{Fan2022CGCN} & TMM 2022  & ResNet50  & 54.40 & 60.03 & 83.68 & 70.49 & - & - & - & -  \\ 
VTB \citep{cheng2022VTB} & TCSVT 2022  & ViT-B$/$16 & 69.44 & 78.28 & 84.39 & 80.84 & 67.48 & 76.41 & 83.32 & 79.35 \\
PARFormer-L \citep{fan2023parformer} & TCSVT 2023 & Swin-L & 69.94 & 79.63 & \textcolor{red}{\textbf{88.19}} & 81.35 & - & - & - & - \\
DFDT \citep{zheng2023diverse} & EAAI 2023 & Swin-B & \textcolor{blue}{\textbf{70.89}}  & 80.36 & 84.32 & \textcolor{blue}{\textbf{82.15}} & \textcolor{blue}{\textbf{69.30}} & \textcolor{blue}{\textbf{79.38}} & 82.62 & \textcolor{blue}{\textbf{80.97}} \\
OAGCN \citep{lu2023orientation} & TMM 2023 & Swin-B & 69.32 & 78.32 & \textcolor{blue}{\textbf{87.29}} & \textcolor{red}{\textbf{82.56}} & - & - & - & - \\
VTB* \citep{cheng2022VTB} & TCSVT 2022  & ViT-L$/$14  & 69.78 & 78.09 & 85.21 & 81.10 & 67.58 & 76.19 & \textcolor{red}{\textbf{84.00}} & 79.52 \\	
{SSPNet}~\citep{shen2024sspnet} & {PR 2024} & Swin-S & 70.21 & 80.14 & 82.90 & 81.50 & - & - & - & - \\
{ViT-RE++}~\citep{tan2024vision} & {TMM 2024} & ViT-B & 69.45 & 81.18 & 80.80 & 80.99 & - & - & - & - \\
{SOFA}~\citep{wu2024selective} & {AAAI 2024} & ViT-B & 70.03 & 79.99 & 83.03 & 81.15 & 68.62 & 78.00 & 83.14 & 80.15 \\
{SOFAFormer++}~\citep{wu2025learning} & {TCSVT 2025} & ViT-B & 69.73 & 78.77 & 84.05 & 80.97 & 67.86 & 77.57 & \textcolor{blue}{\textbf{83.81}} & 79.66 \\
{HDFL}~\citep{wu2025high} & {NN 2025} & ViT-B & 70.64 & 80.94 & 82.82 & 81.55 & - & - & - & - \\
{AAR}~\citep{wu2025rethinking} & {NC 2025} & ViT-B & 70.49 & 80.25 & 83.55 & 81.51 & 68.22 & 78.30 & 82.18 & 79.85 \\
{PIL}~\citep{zhou2025solution} & {IJCV 2025} &  ConvNeXt-base & - & - & - & 80.47 & - & - & - & - \\
\hline

SequencePAR & - & ViT-L$/$14 & \textcolor{red}{\textbf{71.47}} & \textcolor{red}{\textbf{82.40}} & 82.09 & 82.05 & \textcolor{red}{\textbf{70.14}} & \textcolor{red}{\textbf{81.37}} & 81.22 & \textcolor{red}{\textbf{81.10}}\\

\hline \toprule [0.5 pt] 
\end{tabular} } 
\end{table*}

\begin{table*}[!htb]
\center
\small   
\caption{Comparison with state-of-the-art methods on PETA-ZS and RAP-ZS datasets.} \label{PARZSresults} 
\resizebox{1\columnwidth}{!}{
\begin{tabular}{l|c|c|cccc|cccc}
\hline \toprule [0.5 pt] 
\multicolumn{1}{c|}{\multirow{2}{*}{Methods}} & \multicolumn{1}{c|}{\multirow{2}{*}{Ref}} & \multicolumn{1}{c|}{\multirow{2}{*}{Backbone}} & \multicolumn{4}{c|}{PETA-ZS} & \multicolumn{4}{c}{RAP-ZS} \\ \cline{4-11}
\multicolumn{1}{c|}{} &
  \multicolumn{1}{c|}{} &
  \multicolumn{1}{c|}{} &
  \multicolumn{1}{c}{Accuracy} & 
  \multicolumn{1}{c}{Precision} &
  \multicolumn{1}{c}{Recall} &
  \multicolumn{1}{c|}{F1} &
  \multicolumn{1}{c}{Accuracy} &
  \multicolumn{1}{c}{Precision} &
  \multicolumn{1}{c}{Recall} &
  \multicolumn{1}{c}{F1} \\ \hline
VAC\citep{2019VAC} & CVPR 2019  & ResNet50 & 57.72 & 72.05 & 70.64 & 70.90 & 63.25 & 76.23 & 76.97 & 76.12 \\
Jia et al.\citep{2021Rethinking} & -  & ResNet50 & 58.19 & 73.09 & 70.33 & 71.68 & 63.61 & 76.88 & 76.62 & 76.75 \\
VTB \citep{cheng2022VTB} & TCSVT 2022 & ViT-B$/$16 & 60.50 & 73.29 & 74.40 & 73.38  & 64.73 & 74.93 & \textcolor{blue}{\textbf{80.85}} & 77.35 \\
VTB*\citep{cheng2022VTB} & TCSVT 2022  & ViT-L$/$14  & \textcolor{blue}{\textbf{63.12}} & 74.77 & \textcolor{blue}{\textbf{77.24}} & 75.50  & \textcolor{blue}{\textbf{68.34}} & 76.81 & \textcolor{red}{\textbf{84.51}} & \textcolor{blue}{\textbf{80.07}} \\
{SOFA}~\citep{wu2024selective} & {AAAI 2024} & ViT-B & 62.07 & 74.97 & 75.13 & 74.63 & 66.26 & 78.20 & 79.44 & 78.42 \\
{SOFAFormer++}~\citep{wu2025learning} & {TCSVT 2025} & ViT-B & 61.41 & 74.09 & 75.73 & 74.10 & 66.09 & 77.53 & 78.33 & 78.33 \\
{HDFL}~\citep{wu2025high} & {NN 2025} & ViT-B & 62.01 & 75.07 & 75.36 & 74.78 & 66.70 & 78.52 & 79.81 & 78.42 \\
{AAR}~\citep{wu2025rethinking} & {NC 2025} & ViT-B & 62.89 & \textcolor{blue}{\textbf{75.93}} & 75.66 & 75.38 & 66.51 & \textcolor{blue}{\textbf{78.83}} & 79.13 & 78.60 \\
{PIL}~\citep{zhou2025solution} & {IJCV 2025} & ConvNeXt-base & - & - & - & \textcolor{blue}{\textbf{75.81}} & - & - & - & 79.75 \\

\hline
SequencePAR & - & ViT-L$/$14  & \textcolor{red}{\textbf{66.70}}  & \textcolor{red}{\textbf{78.75}}  & \textcolor{red}{\textbf{78.52}}  & \textcolor{red}{\textbf{78.40}}         & \textcolor{red}{\textbf{70.28}} & \textcolor{red}{\textbf{82.13}} & 80.55 & \textcolor{red}{\textbf{81.14}}  \\
\hline \toprule [0.5 pt] 
\end{tabular} 
}
\end{table*}

\subsection{Implementation Details} \label{impleDetails} 
In our experiments, we adopt the ViT-L$/$14 version of the CLIP~\citep{radford2021CLIP} model as our feature extractor. The vision encoder of our model is a 24-layer vision Transformer~\citep{dosovitskiy2020VIT} network and the dimension of the hidden layer is 1024. Our sequence generation network is a 6-layer Transformer Decoder, each with eight attention heads. The dimension of the hidden layer is 768. Considering the pedestrian images are all long strip shapes, however, the resolution of the input images for the CLIP is $224 \times 224$. To adapt to the input resolution, we padded the raw pedestrian images using black pixels into a square shape and resized them into $224 \times 224$. Note that, the randomly crop and flip operations are also adopted for data augmentation.

During the training phase, we utilize the ground truth as the input and predict the attributes in a parallel manner, which can reduce the cumulative error significantly and make our model converge faster. For the model inference, we adopt a step-by-step generation approach that uses the output predicted in the previous step and historical output as the inputs of our network. For the detailed parameters, we set the learning rate as 1e-5. We train our model for 50 epochs with the Adam optimizer and set the batch size as 32. 
Our source code is implemented using Python and the deep learning framework PyTorch~\citep{paszke2019pytorch}. The experiments are conducted based on a server with GPU A100. More details can be found in our source code.

\subsection{Comparison with Other SOTA Models} \label{CompSOTA}  

In this section, we will report our recognition results on all six datasets and compare them with existing state-of-the-art pedestrian attribute recognition algorithms. Note that the results of VTB* are obtained by replacing the backbone network of VTB using the visual encoder of the CLIP model. 

\noindent \textbf{Results on PETA dataset.~} 
As shown in Table~\ref{PETAPA100Kresults}, our proposed SequencePAR model achieves the best performance on most of the evaluation metrics, i.e., 84.92, 90.44, 90.73, 90.46 on the Accuracy, Precision, Recall, and F1-measure, respectively. Compared with other Transformer-based pedestrian attribute recognition models, we can find that our model exceeds both the VTB (ViT-base$/$16, 79.60, 86.76, 87.17, 86.71) and VTB* (ViT-L$/$14, 79.59, 86.66, 87.82,  86.97) by +5.32$/$+3.68$/$+3.56$/$+3.67 and +5.33$/$ +3.78$/$+2.91$/$+3.41. Our model also achieves better performance than the DRFormer~\citep{2022drformer} on the Accuracy, Precision, and F1-measure metric. Therefore, we can draw the conclusion that our model achieves state-of-the-art results on the PETA dataset, which fully validates the superiority of our sequence generation framework for the pedestrian attribute recognition task.

\noindent \textbf{Results on PA100K dataset.~} 
As shown in Table~\ref{PETAPA100Kresults}, our model SequencePAR achieves 83.94, 90.38, 90.23, and 90.10 in Accuracy, Precision, Recall, and F1, respectively. These results surpass those of the multi-label classification model VTB* (81.76, 87.87, 90.67, and 88.86), which also uses the ViT-L$/$14 model as the backbone. These experiments demonstrate that our attribute generation-based approach outperforms the discriminative-based learning framework. Additionally, our model achieves superior results compared to the state-of-the-art recognition models.

\noindent \textbf{Results on RAPv1 dataset.~}
As shown in Table~\ref{RAPResult}, our model achieves 71.47, 82.40, 82.09, and 82.02 in accuracy, precision, recall, and F1, respectively. In contrast, the strong baseline VTB*, a discriminative method, achieves 69.78, 78.09, 85.21, and 81.10 on these metrics. Our results clearly outperform those of the baseline and other compared attribute recognition models, including DAFL~\citep{jia2022learning} (AAAI-2022) and DRFormer~\citep{2022drformer} (NC 2022). These experiments comprehensively validate the effectiveness of our proposed model for pedestrian attribute recognition.

\noindent \textbf{Results on RAPv2 dataset.~}
As shown in Table~\ref{RAPResult}, the multi-label classification-based attribute recognition model VTB* (backbone: ViT-L$/$14) achieves 67.58, 76.19, 84.00, and 79.52 on the four metrics, respectively. In comparison, our model achieves 70.14, 81.37, 81.22, and 81.10. Therefore, we can conclude that our model outperforms the strong discriminative learning-based pedestrian attribute recognition algorithm.

\noindent \textbf{Results on PETA-ZS dataset.~} 
In addition to the aforementioned standard evaluation, we also conduct experiments on zero-shot setting datasets in this work. According to the experimental results reported in Table~\ref{PARZSresults}, our model achieves 66.70, 78.75, 78.52, and 78.40 in Accuracy, Precision, Recall, and F1, respectively. These results outperform those of the multi-label classification model VTB* (63.12, 74.77, 77.24, 75.50), which also uses the ViT-L$/$14 model as the backbone network.

\noindent \textbf{Results on RAP-ZS dataset.~} 
The experimental results reported on the RAP-ZS dataset, as illustrated in Table~\ref{PARZSresults}, show that the baseline method VTB* achieves 68.34, 76.81, 84.51, and 80.07 in Accuracy, Precision, Recall, and F1, respectively. However, our model achieves better results on these metrics, namely 70.28, 82.13, 80.55, and 81.14. Experiments on the PETA-ZS and RAP-ZS datasets also show that our model performs better in this setting. We attribute this to the use of a large visual-language model, which exhibits better generalization ability. Another key reason is that the generative learning-based strategy is more effective for pedestrian attribute recognition.

\begin{table}[!htp]
\center
\small  
\caption{Compare the greedy search and beam search with different beam widths. The best results are highlighted in \textbf{bold}. } \label{BeamSearch} 
\begin{tabular}{c|cccc}
\hline \toprule [0.5 pt]
\multicolumn{1}{c|}{\multirow{2}{*}{Beam Width}} & \multicolumn{4}{c}{PETA}  \\ \cline{2-5}
\multicolumn{1}{c|}{} &
  \multicolumn{1}{c}{Accuracy} &
  \multicolumn{1}{c}{Precision} &
  \multicolumn{1}{c}{Recall} &
  \multicolumn{1}{c}{F1} \\ \hline 
\textbf{1} & \textbf{84.92}  & \textbf{90.44}  &  90.73 & \textbf{90.46}  \\
3 & 84.90 & 90.33 & 90.79 & 90.45 \\
5 & 84.89 & 90.36 & 90.76 & 90.44 \\
10 & 84.84 & 90.27 & \textbf{90.80} & 90.41 \\
\hline \toprule [0.5 pt] 
\end{tabular} 
\end{table}

\subsection{Ablation Study} \label{ablationStudy}

In this work, we conduct extensive experiments to help the readers better understand the interesting factors related to our proposed SequencePAR model.

\noindent \textbf{Greedy Search or Beam Search for Attribute Generation? }
For the sequence generation tasks, like machine translation and image$/$video captioning, the greedy search and beam search are all widely used in the inference phase. The greedy search policy will select and remain the element with the maximum response score. In contrast, the beam search policy will always select and retain a fixed number of candidates (also called beam width $\mathcal{B}$) for each time step, and select the trajectory with the maximum summarised response.

In this part, we evaluate different decoding strategies to verify their effectiveness for pedestrian attribute generation. As shown in Table~\ref{BeamSearch}, the greedy search achieves $84.74$, $90.38$, $90.42$, and $90.28$ on Accuracy, Precision, Recall, and F1, respectively. When beam search is adopted, the recognition performance on the PETA dataset remains comparable to that of greedy search, but no further improvement is observed. Therefore, we employ the simple greedy search strategy for all subsequent experiments in this work.

\noindent \textbf{Results using Different Layers of Decoders. }
As shown in Table~\ref{decoderLayers}, we conduct experiments on the PETA dataset to examine the impact of varying the number of layers in our decoder network. Specifically, we set the number of layers to 1, 3, 6, 9, and 12. Table~\ref{decoderLayers} shows that the F1 scores are 89.80, 89.68, 90.46, 89.60, and 89.76, demonstrating relative stability. The best recognition performance on the PETA dataset is achieved when six decoding layers are used. Therefore, we use six layers as the default configuration for our decoder network in subsequent experiments.

\begin{table}[!htp]
\center
\small  
\caption{Comparison of Different Decoding Layers on PETA dataset. The best results are highlighted in \textbf{bold}.}  
\label{decoderLayers} 
\begin{tabular}{c|cccc}
\hline \toprule [0.5 pt]
\multicolumn{1}{c|}{\multirow{2}{*}{Decoder Layers}} & \multicolumn{4}{c}{PETA}  \\ \cline{2-5}
\multicolumn{1}{c|}{} &
  \multicolumn{1}{c}{Accuracy} &
  \multicolumn{1}{c}{Precision} &
  \multicolumn{1}{c}{Recall} &
  \multicolumn{1}{c}{F1} \\ \hline 
1 & 83.96 & 89.90 & 89.95 &  89.80\\
3 & 83.84 & 89.82 & 89.78 & 89.68 \\
\textbf{6} & \textbf{84.92}  & \textbf{90.44}  &  \textbf{90.73} & \textbf{90.46}  \\
9 & 83.72 & 89.67 &89.78 & 89.60 \\
12 & 83.90 & 89.93 & 89.84 & 89.76\\
\hline \toprule [0.5 pt] 
\end{tabular} 
\end{table}

\begin{figure*}[!htp]
\centering
\small
\includegraphics[width=\linewidth]{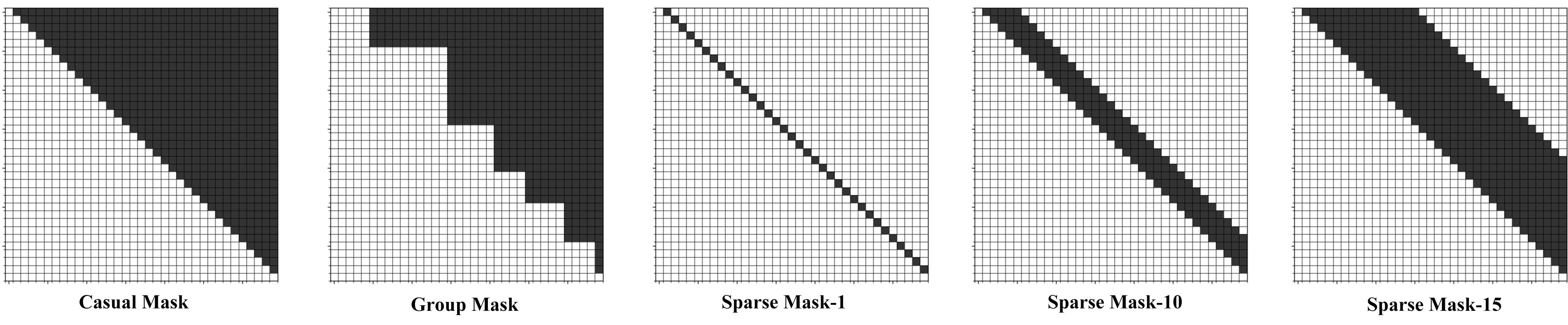}
\caption{Visualization of Different Designs of The Masking Strategy. \textcolor{gray}{Gray} represents the invisible.} 
\label{mask}
\end{figure*}

\noindent {\textbf{Effectiveness of Causal Masks.}}
{As shown in Table~\ref{masking}, we first analyze the necessity of causal masking to determine the most effective masking strategy. We observe that removing the attention mask from the autoregressive decoder causes model failure. This is because we use the true values as input to reduce cumulative error during training, enabling the model to complete the entire prediction in one forward pass. Without the attention mask, the model can anticipate predictions through the attention mechanism, causing training to collapse. We then experiment with several masking strategies, including the "group masking" strategy, where groups of attributes (e.g., head, top dress, bottom dress, etc.) are visible to each other, and causal masks are applied between attribute groups to model relationships. Sparse masking means the model cannot see the next K tokens. Experiments show that the optimal performance is achieved using the causal masking strategy. Specifically, the group masking strategy underperforms because elements within attribute groups can see future tokens, causing overfitting and significant performance degradation. Additionally, the sparse masking strategy generally performs poorly, especially with smaller mask step settings (e.g., Sparse Mask-1). Smaller mask lengths do not effectively obscure future tokens, leading to information leakage and causing the model to depend on future information, which negatively impacts training. However, increasing the mask length (e.g., Sparse Mask-10) improves performance. This is because, for attributes appearing later in the sequence, a larger mask length effectively obscures future tokens, leading to improved predictions. In summary, our results underscore the importance of the causal masking strategy. Causal masking prevents future information leakage, enhancing the model’s generalization ability.}

\begin{table}[!htp]
\center
\small  
\caption{Comparison of Different Designs of The Masking Strategy on PETA dataset. }  
\label{masking} 
\begin{tabular}{c|cccc}
\hline \toprule [0.5 pt]
\multicolumn{1}{c|}{\multirow{2}{*}{Masking Strategy}} & \multicolumn{4}{c}{PETA}  \\ \cline{2-5}
\multicolumn{1}{c|}{} &
  \multicolumn{1}{c}{Accuracy} &
  \multicolumn{1}{c}{Precision} &
  \multicolumn{1}{c}{Recall} &
  \multicolumn{1}{c}{F1} \\ \hline 
Causal Mask & \textbf{84.92}  & \textbf{90.44}  &  \textbf{90.73} & \textbf{90.46}  \\
Group Mask & 64.09 & 88.51 & 68.13 & 76.71 \\
Sparse Mask-1 & 8.72 & 37.71 & 9.98 & 15.67\\
Sparse Mask-10 & 17.74 & 72.53 & 18.05 & 27.73\\
Sparse Mask-15 & 22.07 & 84.91 & 22.42 & 34.29\\
w/o Mask & 7.18 & 37.33 & 7.52 & 12.28 \\
\hline \toprule [0.5 pt] 
\end{tabular} 
\end{table}

\begin{figure}[!htp]
\centering
\small
\includegraphics[width=0.8\linewidth]{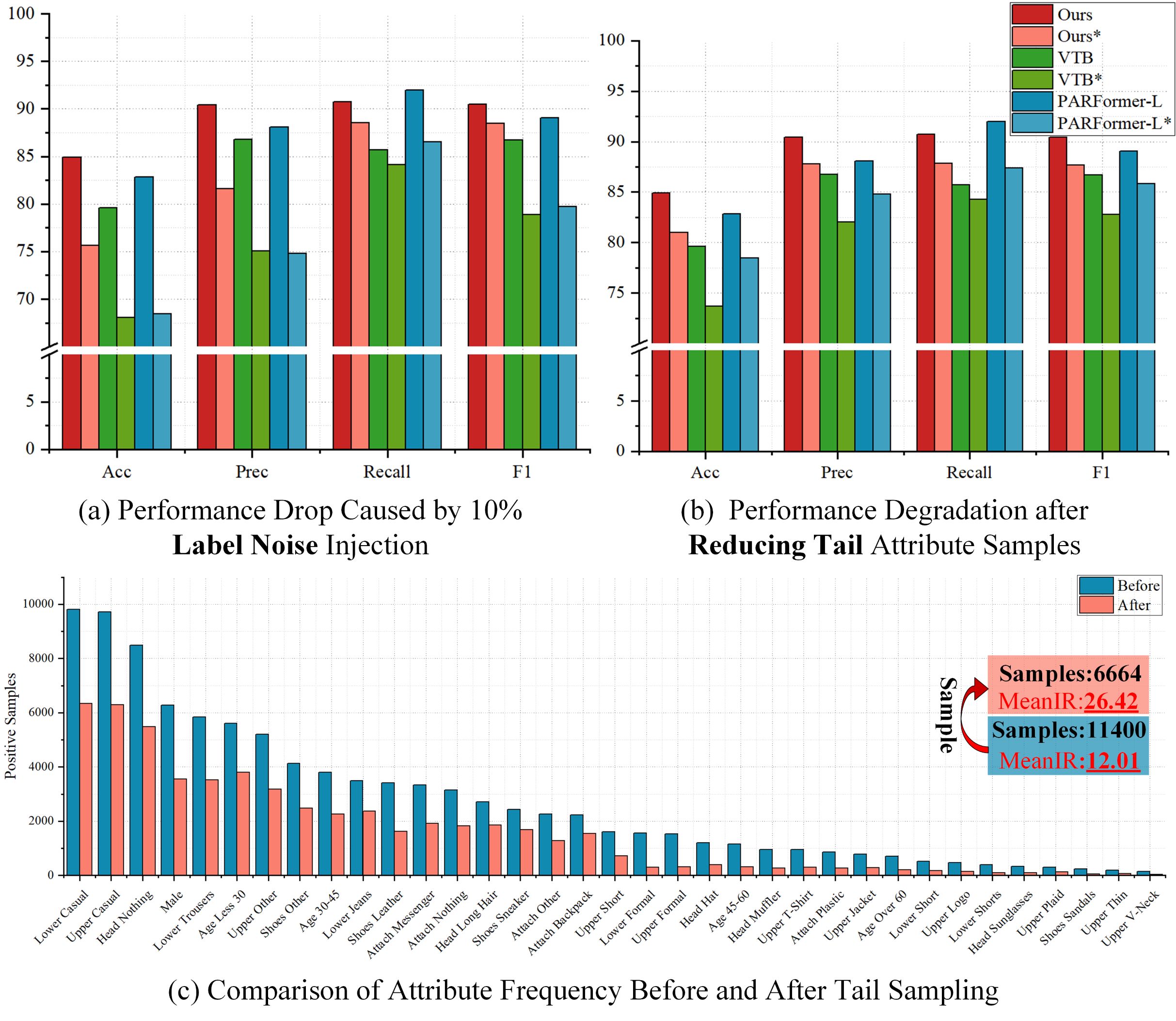}
\caption{Robustness Analysis under Label Noise and Long-tailed Distribution.
(a) Performance degradation under 10\% random label noise.
(b) Performance after reducing tail attribute samples.
(c) Attribute frequency distribution after reducing tail sampling. } 
\label{NoiseTailed}
\end{figure}

\begin{table}[!htp]
\center
\small  
\caption{Comparison of Different Prompts for Attributes Expanding on PETA dataset. }  
\label{prompts} 
\begin{tabular}{c|cccc}
\hline \toprule [0.5 pt]
\multicolumn{1}{c|}{\multirow{2}{*}{Prompts}} & \multicolumn{4}{c}{PETA}  \\ \cline{2-5}
\multicolumn{1}{c|}{} &
  \multicolumn{1}{c}{Accuracy} &
  \multicolumn{1}{c}{Precision} &
  \multicolumn{1}{c}{Recall} &
  \multicolumn{1}{c}{F1} \\ \hline 
$<$CLASS$>$ & 84.23 & 89.88 & 90.32 & 89.98 \\
A photo of a $<$CLASS$>$ & 84.53 & 90.15 & 90.49 & 90.19 \\
Custom Template & \textbf{84.92}  & \textbf{90.44}  &  \textbf{90.73} & \textbf{90.46}  \\
\hline \toprule [0.5 pt] 
\end{tabular} 
\end{table}

\begin{table}
\center
\small  
\caption{Comparison of Different orders for Expanding on the PETA dataset. }  
\label{Order} 
\begin{tabular}{c|cccc}
\hline \toprule [0.5 pt]
\multicolumn{1}{c|}{\multirow{2}{*}{Prompts}} & \multicolumn{4}{c}{PETA}  \\ \cline{2-5}
\multicolumn{1}{c|}{} &
  \multicolumn{1}{c}{Accuracy} &
  \multicolumn{1}{c}{Precision} &
  \multicolumn{1}{c}{Recall} &
  \multicolumn{1}{c}{F1} \\ \hline 
Init & 84.92 & \textbf{90.44} & 90.73 & 90.46  \\
Inverse & \textbf{85.01}  & 90.41  &  \textbf{90.89} & \textbf{90.53}  \\
Shuffle 1  & 84.69 & 90.23 & 90.49 & 90.25  \\
Shuffle 1  & 84.62 & 90.25 & 90.42 & 90.21  \\
\hline \toprule [0.5 pt] 
\end{tabular} 
\end{table}

\noindent {\textbf{Effectiveness of Prompts for Attributes Expanding.}}
As shown in Table~\ref{prompts}, we explored the effect of using two additional cues on the model's performance. The template we used was more detailed and contextually appropriate, e.g., \textit{"the \underline{age} of this pedestrian is \underline{less than 40} years old"}. Additionally, we explored two alternative cues. The first approach involved using the attribute words themselves directly. The results indicate a slight decrease in model performance across the four evaluation metrics (0.69/0.56/0.41/0.48), demonstrating that the cue template enhances attribute differentiation and mitigates the expression ambiguity of attribute words. The second approach involved using the commonly adopted CLIP template, \textit{"A photo of a $<$CLASS$>$"}. The results show a slight decline in performance (0.39/0.29/0.24/0.27 for Acc/Prec/Recall/F1, respectively), further confirming that an appropriate semantic template aids the model in better understanding attribute semantics.

\begin{table}
\center
\small  
\caption{Comparison of Model Performance under Original, Noisy, and Long-tailed Training Settings.}  
\label{Training} 
\resizebox{0.8\columnwidth}{!}{
\begin{tabular}{c|l|llll}
\hline \toprule [0.5 pt]
\multicolumn{1}{c|}{\multirow{2}{*}{Train Set}} &  \multicolumn{1}{c|}{\multirow{2}{*}{Methods}} &\multicolumn{4}{c}{PETA}  \\ \cline{3-6}
\multicolumn{1}{c|}{} &
\multicolumn{1}{c|}{} &
  \multicolumn{1}{l}{Accuracy} &
  \multicolumn{1}{l}{Precision} &
  \multicolumn{1}{l}{Recall} &
  \multicolumn{1}{l}{F1} \\ \hline 
\multirow{3}{*}{Origin} & Ours & 84.92 & 90.44 & 90.73 & 90.46  \\
& VTB\citep{cheng2022VTB} & 79.60 & 86.76 & 85.69 & 86.71  \\
& PARFormer-L\citep{fan2023parformer} & 82.86 & 88.06 & 91.98 & 89.06  \\ \hline

\multirow{3}{*}{10\% Noise} & Ours & 75.66\textbf{\textcolor{red}{$\downarrow$9.26}} & 81.60\textbf{\textcolor{red}{$\downarrow$8.84}} & 88.55\textcolor{DarkRed}{$\downarrow$2.18} & 84.55\textbf{\textcolor{red}{$\downarrow$5.19 }} \\
& VTB\citep{cheng2022VTB} & 68.06\textcolor{DarkRed}{$\downarrow$11.54} & 75.07\textcolor{DarkRed}{$\downarrow$11.69} & 84.15\textbf{\textcolor{red}{$\downarrow$1.54}} & 
 78.85\textcolor{DarkRed}{$\downarrow$10.85} \\
& PARFormer-L\citep{fan2023parformer} & 68.44\textcolor{DarkRed}{$\downarrow$14.42} & 74.82\textcolor{DarkRed}{$\downarrow$12.62} & 86.50\textcolor{DarkRed}{$\downarrow$5.48} & 79.69\textcolor{DarkRed}{$\downarrow$9.37}  \\ \hline

\multirow{3}{*}{Long-tailed} & Ours & 80.97\textbf{\textcolor{red}{$\downarrow$3.95}} & 87.79\textbf{\textcolor{red}{$\downarrow$2.65}} & 87.85\textcolor{DarkRed}{$\downarrow$2.88} & 87.68\textbf{\textcolor{red}{$\downarrow$2.78}} \\
& VTB\citep{cheng2022VTB} & 73.70\textcolor{DarkRed}{$\downarrow$5.90} & 82.02\textcolor{DarkRed}{$\downarrow$4.74} & 84.29\textbf{\textcolor{red}{$\downarrow$1.40}} & 
82.80\textcolor{DarkRed}{$\downarrow$3.91} \\
& PARFormer-L\citep{fan2023parformer} & 78.47\textcolor{DarkRed}{$\downarrow$4.39} & 84.81\textcolor{DarkRed}{$\downarrow$3.25} & 87.36\textcolor{DarkRed}{$\downarrow$4.62} & 85.82\textcolor{DarkRed}{$\downarrow$3.24}  \\ \hline

\hline \toprule [0.5 pt] 
\end{tabular}}
\end{table}

\noindent {\textbf{Analysing the Impact of Different Attribute Orders.}}
To analyze the sequence relationships between attributes learned by the model, we randomly disrupted the PETA attribute list twice and compared the results with the original order. As shown in Table~\ref{Order}, the model still effectively captures the relationships between attributes with minimal deviation, even when their order is altered. However, disrupting the order led to a slight decrease in performance, likely due to the causal relationships between certain attributes. Across all metrics, performance dropped by approximately 0.2\%. These results indicate that while the model demonstrates strong robustness to changes in attribute order, alterations to certain causal relationships still impact performance.

\begin{figure}[!htp]
\centering
\small
\includegraphics[width=0.5\linewidth]{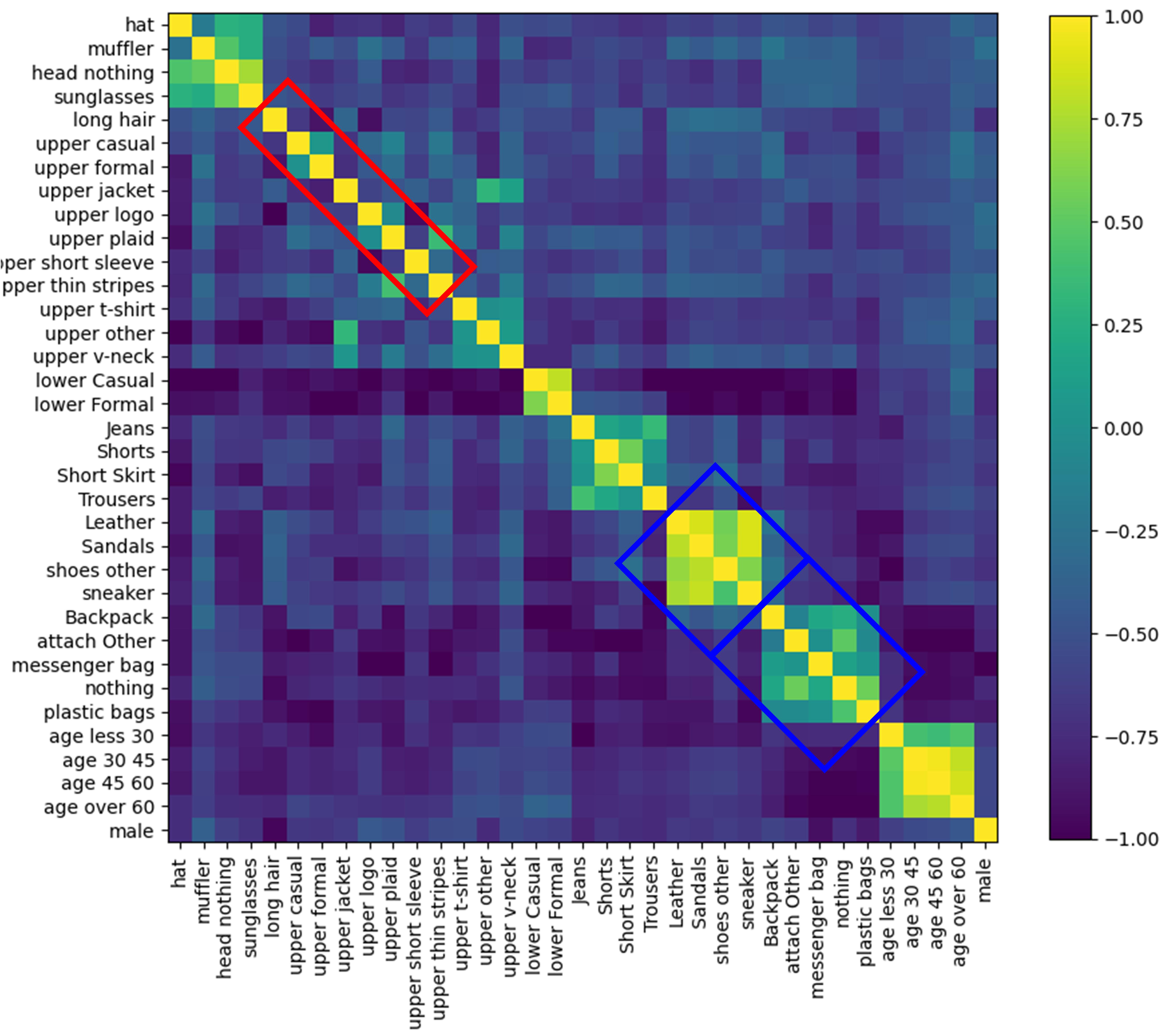}
\caption{Visualization of the similarity matrix of pedestrian attributes on the PETA dataset obtained by our SequencePAR model. } 
\label{similarityMatrix}
\end{figure}

\noindent \textbf{Robustness Analysis under Label Noise and Long-tailed Distribution.}
As shown in Figure~\ref{NoiseTailed}. To evaluate the robustness of our model against noisy annotations and imbalanced data, we conduct experiments involving random noise injection and long-tail sampling on the training dataset.

To assess the model’s ability to handle label noise, we introduce 10\% label corruption into the training set. Specifically, each ground-truth attribute label was flipped with a probability of 10\%, resulting in a noisy training set denoted as $D_{noise}$. We then retrain our model, the visual-text fusion method VTB, and the prior state-of-the-art method PARFormer on $D_{noise}$. As shown in Table~\ref{Training}. Under this noise setting, our model shows smaller performance drops across key metrics (Accuracy/Precision/Recall/F1: 9.26/8.84/2.18/5.39) compared to VTB (11.54/11.69/1.54/10.85). Although VTB exhibits a slightly smaller drop in Recall, our model maintains more balanced and stable Precision and Recall, resulting in superior overall F1 performance. Furthermore, compared to PARFormer~\citep{fan2023parformer}, our model experiences smaller drops (5.16/3.78/3.30/2.18), highlighting the enhanced robustness provided by the generative framework. We attribute this to the autoregressive decoder, which models inter-attribute dependencies by conditioning each prediction on previously generated tokens, thus promoting context-aware denoising.

To further investigate robustness under extreme class imbalance, we construct a more severely long-tailed training set through attribute-level sampling. For each attribute, we identify tail classes with relatively few positive samples and randomly remove a fixed proportion of their positive labels. This increases the dataset’s mean imbalance ratio (Mean IR)~\citep{cui2019class}, a commonly used metric in long-tailed classification, where a higher value indicates greater imbalance. Through this sampling strategy, we increase the Mean IR from 12.01 to 26.42 as illustrated in Figure~\ref{NoiseTailed}, yielding a new long-tailed dataset denoted as $D_{lt}$.

All models exhibit performance degradation when trained on $D_{lt}$, however, our method demonstrates the smallest performance drop. As shown in Table~\ref{Training}, in terms of F1 score, SequencePAR drops only 2.78, while VTB and PARFormer-L drop 3.91 and 3.24, respectively. These results confirm the robustness and generalization capability of our method under both noisy and imbalanced training conditions.

\subsection{Visualization} \label{visualization}

In this section, we provide a visualization of the similarity matrix for the learned pedestrian attributes. Additionally, we present the predicted pedestrian attributes generated by our proposed SequencePAR model.

\begin{figure}[!htp]
\centering
\small
\includegraphics[width=0.8\linewidth]{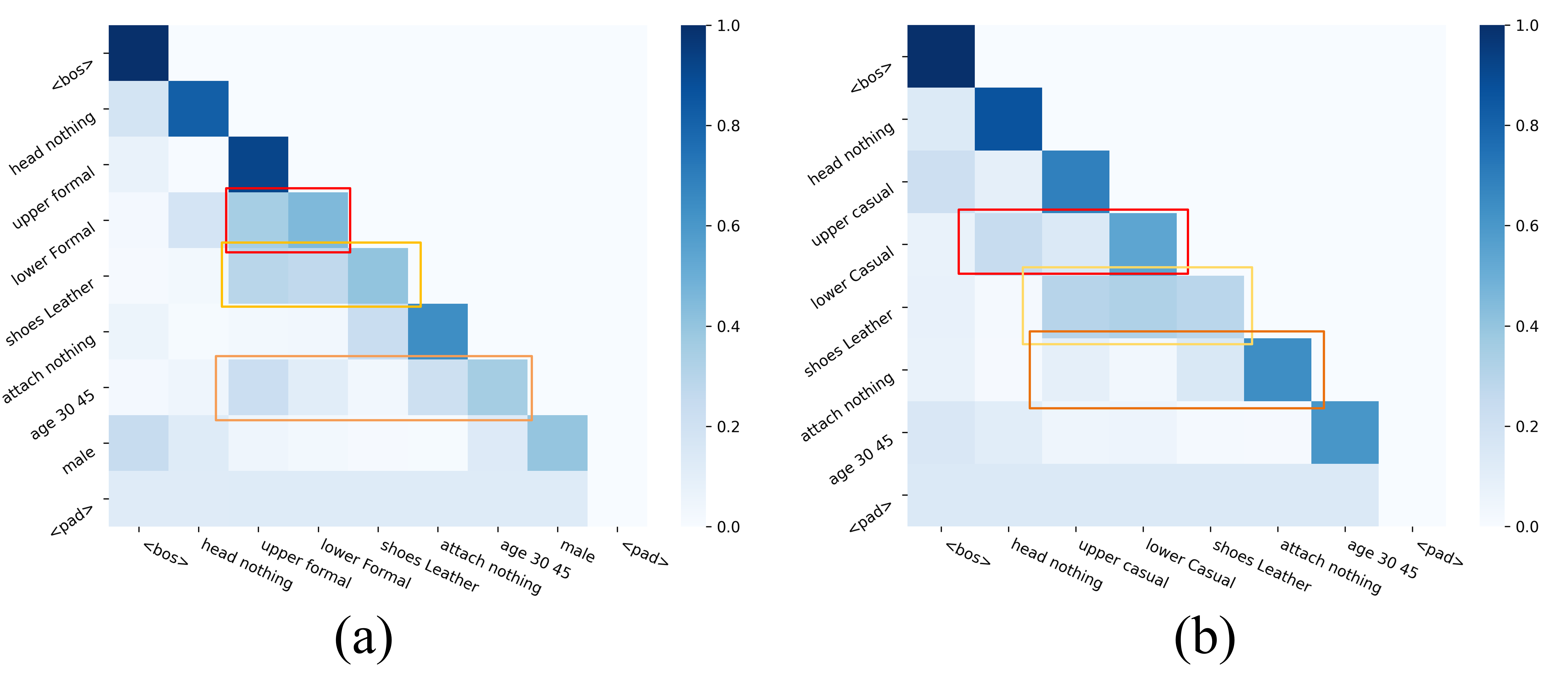}
\caption{Visualization of the attentions between attribute words during our model autoregression.} 
\label{Person}
\end{figure}

\noindent \textbf{Similarity Matrix.~} 
As shown in Fig~\ref{similarityMatrix}, we show the cosine similarity between various pedestrian attributes defined in the PETA datasets. We can find that some attributes are relatively independent, such as long hair, upper casual, upper formal, and upper jacket, as shown in the red rectangle. For attributes like age and shoes, attributes defined in the head region are highly correlated, as shown in blue rectangles. In our future work, we will consider learning the human attributes from the highly correlated groups.

\noindent {\textbf{Attention Map Between Attribute Words.~}}
We also present a visualization of the model's attention during autoregression. As shown in Figure~\ref{Person}(a), the red box highlights the model's increased attention to the \textit{Upper Formal} attribute when predicting the \textit{Lower Formal} attribute. Indeed, a causal relationship exists between these two attributes, and the model successfully captured this connection. Additionally, as shown in the yellow area of Figure~\ref{Person}(a), the model attends to both upper and lower body wear when predicting the \textit{Shoes Leather} attribute, further demonstrating the effective utilization of contextual information by the model.

\begin{figure}[!htp]
\centering
\small
\includegraphics[angle=90,width=0.8\linewidth]{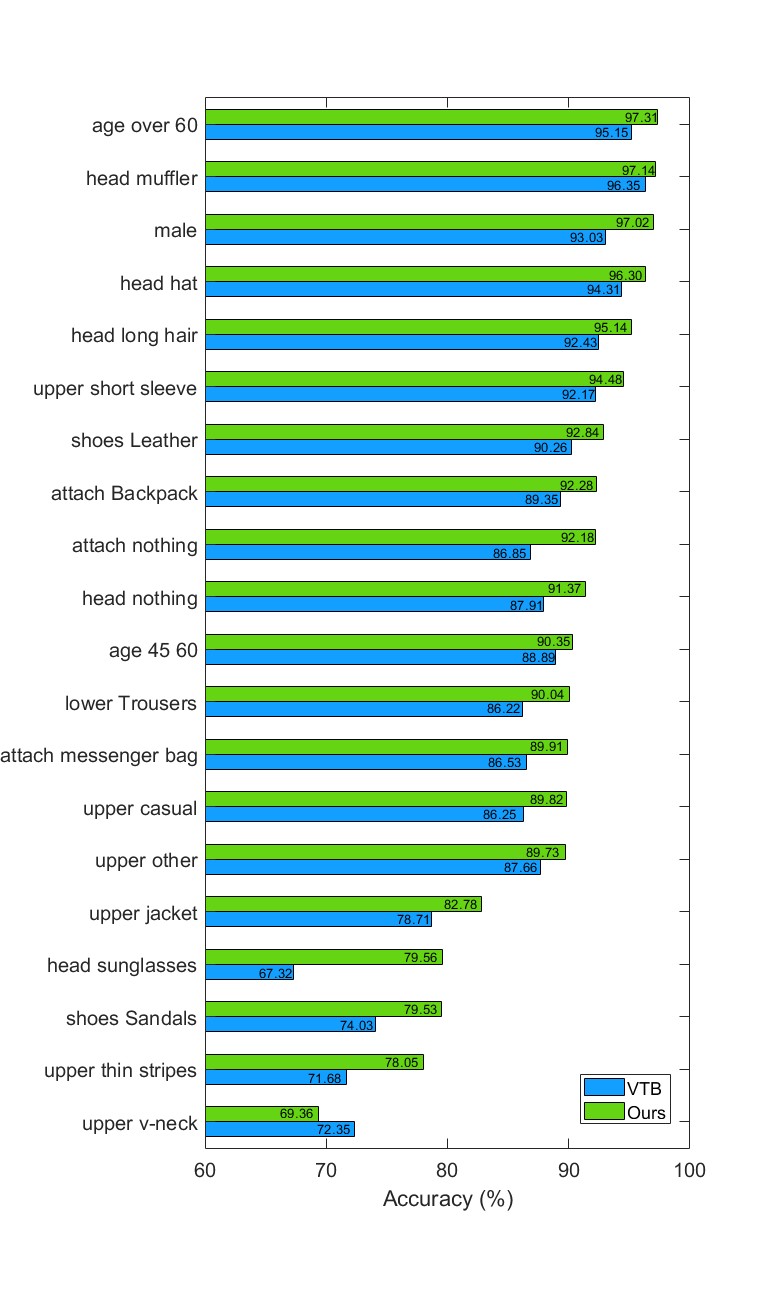}
\caption{Average accuracy of attributes on the PETA dataset.} 
\label{meanAcc20attributes}
\end{figure}

\noindent \textbf{Average Accuracy of Pedestrian Attribute.~} 
As shown in Figure~\ref{meanAcc20attributes}, we compare the average accuracy of our method with VTB~\citep{cheng2022VTB} on 20 pedestrian attributes on the PETA dataset. Our proposed method performs better than VTB~\citep{cheng2022VTB} on most attributes, and some of the characteristics show significant improvement. 
Examples of these enhancements include ``head sunglasses", ``upper jackets", and ``shoe sandals", which can improve by up to +12\%. These results demonstrate the effectiveness of our generative model for pedestrian attribute recognition.

\begin{figure*}
\centering
\small
\includegraphics[width=0.8\linewidth]{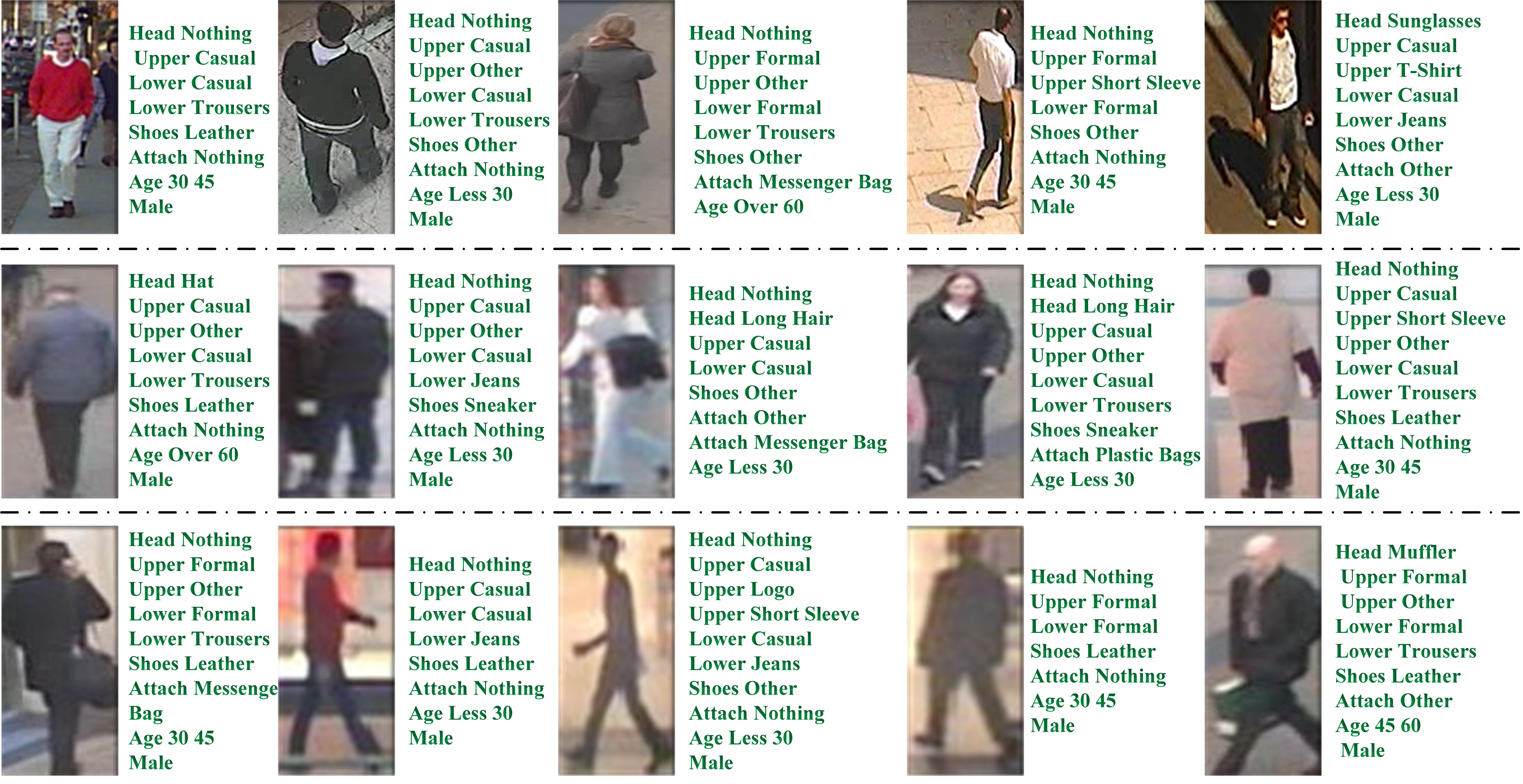}
\caption{Visualization of the predicted pedestrian attributes using our proposed SequencePAR.} 
\label{attResults_vis}
\end{figure*}

\noindent \textbf{Pedestrian Attribute Recognition Results.~} 
As shown in Fig~\ref{attResults_vis}, we give a visualization of predicted attributes of 15 pedestrian images on the PETA dataset. It is evident from our research that our SequencePAR framework accurately predicts attributes such as age, gender, carried items, clothing, etc. These visualizations fully validate the effectiveness of our generative model for the pedestrian attribute recognition problem. 

\begin{figure*}[!htp]
\centering
\small
\includegraphics[width=0.8\linewidth]{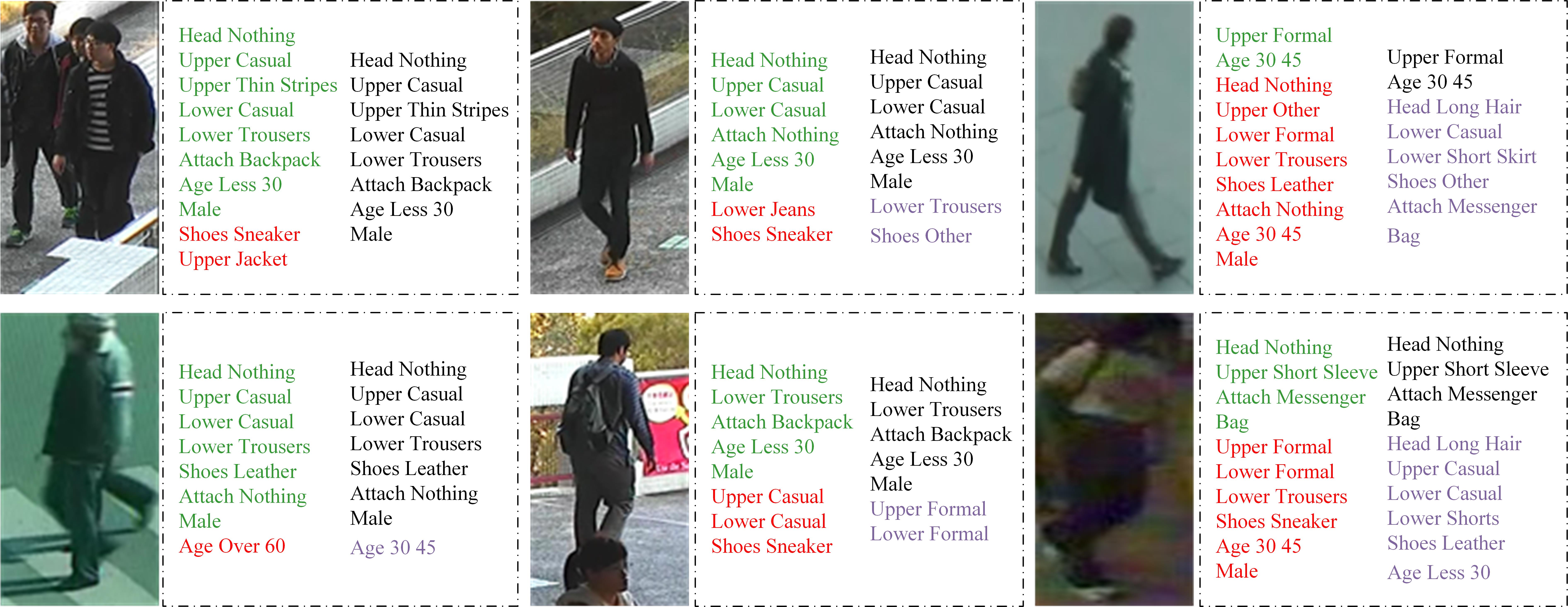}
\caption{Visualization of some incorrectly predicted pedestrian attributes (highlighted in \textcolor{red}{red}) using our proposed SequencePAR.} 
\label{label_error}
\end{figure*}

\begin{figure*}[!htp]
\centering
\small
\includegraphics[width=\linewidth]{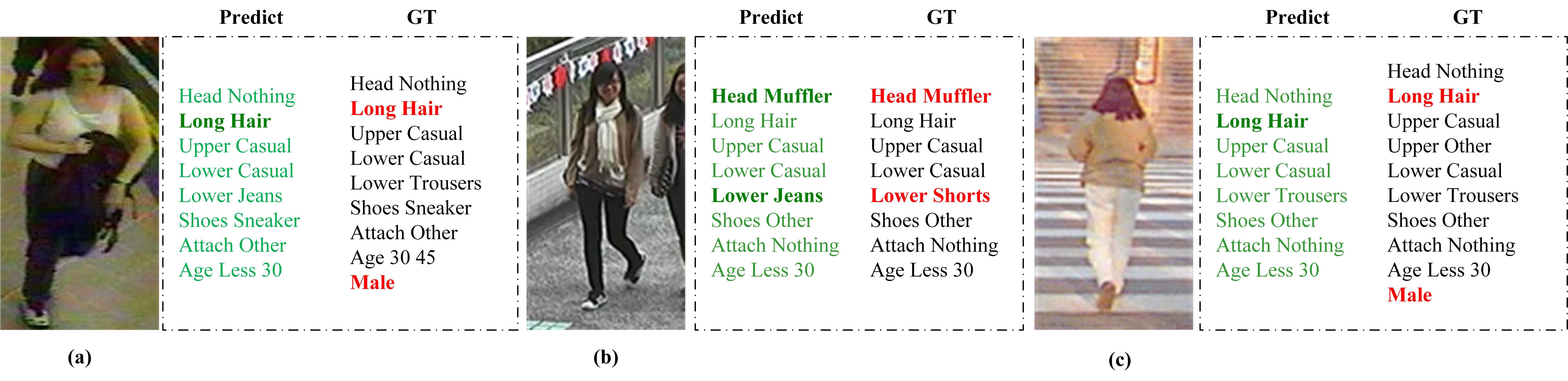}
\caption{Visualization of some incorrectly predicted pedestrian attributes (The counterfactual attributes in the ground truth are represented in \textcolor{red}{red}, and the corrected predictions from our model are represented in \textbf{bold}.) using our proposed SequencePAR.} 
\label{gt_error}
\end{figure*}

\noindent \textbf{{Incorrectly Results.}~} 
{Although our proposed SequencePAR model demonstrates strong performance, as validated in previous experiments, it still faces challenges in certain scenarios. As illustrated in Figure~\ref{label_error}, some predicted pedestrian attributes are incorrect, highlighted in red. These mispredictions often occur in images containing multiple individuals or interfering objects, which are difficult for the model to distinguish. To address this, one potential solution is to first locate the target pedestrian before predicting their attributes. We present the prediction results for noisy samples from the PETA dataset. As shown in Figures~\ref{gt_error} (a) and (c), the dataset incorrectly labels the pedestrian's gender as male, resulting in a conflicting attribute pair of \textit{Long Hair} and \textit{Male}, which our model successfully corrects. Additionally, the dataset contains other types of noise, such as the contradictory attribute pair \textit{Head Muffler} and \textit{Lower Shorts} in Figure~\ref{gt_error} (b), which violates seasonal consistency. As observed, our model mitigates the impact of such noisy samples by leveraging contextual information.
}

{\subsection{Difference with Existing Methods}}
{
We will discuss the differences between our method and existing methods. We categorize related methods into three types: PAR methods~\citep{cheng2022VTB}, sequential methods~\citep{zhao2018GRL}, and generative methods~\citep{2020m2}.} 

{
First, current mainstream multi-label classification-based~\citep{2019VAC, lu2023orientation} PAR methods primarily treat pedestrian attribute prediction as multiple independent binary classification problems. These methods extract features using a backbone and input them into multiple classifiers to predict the confidence of each attribute. The PAR dataset contains a large number of negative samples, which often leads to low prediction confidence from the classifiers. Compared to the discriminative architectures used in mainstream PAR methods, our approach leverages a generative architecture to transform the independent prediction problem into a joint probability problem, thereby better utilizing the contextual information of attributes to logically predict errors (as shown in Figure~\ref{gt_error}). Additionally, our approach does not rely on the binary classifiers commonly used in PAR but instead predicts the index of attribute words in the vocabulary, alleviating the issue mentioned by Jia et al.~\citep{jia2022learning}, where the prediction confidence lies on the decision boundary as the number of attributes increases.}

Second, RNN-based sequential methods~\citep{wang2017MLrecog} typically employ RNNs or LSTMs to sequentially model the dependencies among pedestrian attributes. However, these approaches still rely on discriminative prediction schemes, making them vulnerable to class imbalance issues. Our method transforms discriminative prediction into the task of predicting the index of pedestrian attributes, utilizing a joint positive sample-weighted NLL loss, which mitigates the issue of class imbalance. Moreover, our approach utilizes an advanced self-attention-based Transformer~\citep{vaswani2017Former} network to model attribute relationships. It excels in parallelism, long-range dependencies, and attention mechanisms, with a more detailed analysis presented in Section~\ref{differenceRNN}.

Finally, compared with other generation-based methods such as image captioning~\citep{2020m2}, which typically employ a Transformer encoder-decoder architecture to perform autoregressive target description via masked Multi-Head Self-Attention (MHSA), our approach introduces several key innovations. Although MHSA has been widely used in tasks like image captioning, it has not yet been applied in pedestrian attribute recognition (PAR) to model the causal dependencies among attributes. Rather than directly adopting MHSA, we design a novel framework comprising an attribute query encoder, a sequence generation decoder, and a closed-vocabulary generation strategy optimized for structured attribute prediction. Experimental comparisons demonstrate that, when using the same ViT-L/14 visual encoder, our method outperforms discriminative models such as VTB* by more than 5\% in accuracy on datasets like PETA. In addition, image captioning typically relies on an open vocabulary, which often requires substantial computational resources and diverse sentence-level annotations. Our model proposes an innovative closed attribute vocabulary, thereby reducing the likelihood of generating irrelevant tokens, accelerating convergence, and eliminating the need for additional sentence-level annotations.

\subsection{Difference with Previous RNN-based PAR} \label{differenceRNN} 
Recurrent Neural Networks (RNN) are widely exploited in pedestrian attribute recognition for feature enhancement~\citep{zhao2018GRL}. To be specific, Zhao et al.~\citep{zhao2018GRL} propose to mine both the semantic and spatial correlations in attribute groups and predict the attributes in a group-by-group manner using the LSTM network. Wang et al.~\citep{wang2017MLrecog} propose a recurrent memorized-attention module for multi-label image classification which is interpretable and contextualized. The model includes a spatial Transformer layer to identify attentional regions from convolutional feature maps and an LSTM network to sequentially predict semantic labeling scores for these regions. These approaches commonly utilize RNNs$/$LSTMs for enhancing features across different human parts, mining attentional regions, or capturing correlations between human attributes.

In this paper, we propose a new Transformer-based generative prediction framework for pedestrian attribute recognition. The difference between our work and existing ones can be summarized as follows:  
1). Existing works usually use the RNN$/$LSTM to model the attribute relations, while we adopt the advanced self-attention-based Transformer network to achieve this goal. Therefore, our model performs better on parallelism, long-range dependencies, attention mechanisms, etc. 
2). Many existing RNN$/$LSTM-based PAR models belong to the discriminative models, meanwhile, our proposed SequencePAR is a new generative framework for the pedestrian attribute recognition task. 
3). Existing works predict the pedestrian attributes using the RNN/ LSTM network which is trained from scratch. In contrast, our SequencePAR can make full use of the off-the-shelf pre-trained Transformer networks (e.g., the vision model CLIP~\citep{radford2021CLIP} and language model BERT~\citep{kenton2019bert}). 
Therefore, our proposed SequencePAR is significantly different from previous recurrent neural network-based attribute recognition models. Also, according to the experimental results reported in sub-section \ref{CompSOTA}, our model also exceeds the RNN-based models on all the compared benchmark datasets.

\section{Conclusion and Future Works}~\label{conclusion}
In this work, we propose a novel sequence generation paradigm for pedestrian attribute recognition, termed SequencePAR. It is proposed based on the fact that existing pedestrian attribute recognition (PAR) algorithms are following multi-label or multi-task learning. That is to say, their performance heavily depends on the specific classification heads and is easily influenced by imbalanced data or noisy samples. Therefore, we propose the SequencePAR which belongs to the generative models. To be specific, it extracts the pedestrian features using a pre-trained CLIP model and embeds the attribute set into query tokens under the guidance of text prompts. Then, a Transformer decoder is proposed to generate the human attributes by incorporating the visual features and attribute query tokens. The masked multi-head attention layer is introduced into the decoder module to prevent the model from remembering the next attribute while making attribute predictions during training. Extensive experiments on multiple widely used PAR datasets fully validated the effectiveness of our proposed SequencePAR. 

{This study reveals an unexpected phenomenon through systematic experimental analysis: in the pedestrian attribute prediction task under the generative paradigm, the performance of the beam search strategy is not significantly superior to that of the greedy search strategy. This phenomenon may arise from our closed-set attribute vocabulary design, which lacks synonyms, thus eliminating the need for the model to explore multiple candidate sequences during decoding. 
In our future work, we aim to advance the paradigm in two key dimensions. First, we will construct an open attribute vocabulary and enhance the model's zero-shot prediction capabilities by integrating knowledge distillation techniques from multimodal large language models (MLLMs). This approach will facilitate generalized reasoning for previously unseen attributes. Second, we will employ Chain-of-Thought (CoT) reasoning techniques to improve the model's interpretability in fine-grained attribute association reasoning and counterfactual causal inference while maintaining robust generalization to open-domain scenarios.}

\bibliographystyle{apalike}
\bibliography{reference}


\end{document}